\documentclass[sn-standardnature,iicol]{sn-jnl}

\jyear{2021}%

\theoremstyle{thmstyleone}%
\theoremstyle{thmstyletwo}%

\theoremstyle{thmstylethree}%

\begin{document}

\title[]{Recent advances in artificial intelligence for retrosynthesis}


\author[1]{\fnm{Zipeng} \sur{Zhong}}\email{zipengzhong@zju.edu,cn}
\author[2]{\fnm{Jie} \sur{Song}}\email{sjie@zju.edu,cn}
\author[2]{\fnm{Zunlei} \sur{Feng}}\email{zunleifeng@zju.edu,cn}
\author[3]{\fnm{Tiantao} \sur{Liu}}\email{liutiant@zju.edu.cn}
\author[1]{\fnm{Lingxiang} \sur{Jia}}\email{lingxiangjia@zju.edu,cn}
\author[1]{\fnm{Shaolun} \sur{Yao}}\email{yaoshaolun@zju.edu,cn}
\author*[3]{\fnm{Tingjun} \sur{Hou}}\email{tingjunhou@zju.edu.cn}
\author*[1]{\fnm{Mingli} \sur{Song}}\email{brooksong@zju.edu.cn}

\affil[1]{\orgdiv{College of Computer Science and Technology}, \orgname{Zhejiang University}, \orgaddress{\postcode{310027}, \state{Zhejiang}, \country{P.R. China}}}

\affil[2]{\orgdiv{School of Software Technology}, \orgname{Zhejiang University}, \orgaddress{\postcode{315048}, \state{Zhejiang}, \country{P.R. China}}}

\affil[3]{\orgdiv{Innovation Institute for Artificial Intelligence in Medicine of Zhejiang University, College of Pharmaceutical Sciences}, \orgname{Zhejiang University}, \orgaddress{\postcode{310058}, \state{Zhejiang}, \country{P.R. China}}}

\abstract{
Retrosynthesis is the cornerstone of organic chemistry, providing chemists in material and drug manufacturing access to poorly available and brand-new molecules. 
Conventional rule-based or expert-based computer-aided synthesis has obvious limitations, such as high labor costs and limited search space. In recent years, dramatic breakthroughs driven by artificial intelligence have revolutionized retrosynthesis. Here we aim to present a comprehensive review of recent advances in AI-based retrosynthesis. For single-step and multi-step retrosynthesis both, we first list their goal and provide a thorough taxonomy of existing methods. Afterwards, we analyze these methods in terms of their mechanism and performance, and introduce popular evaluation metrics for them, in which we also provide a detailed comparison among representative methods on several public datasets. In the next part we introduce popular databases and established platforms for retrosynthesis.  Finally, this review concludes with a discussion about promising research directions in this field. 
}

\keywords{retrosynthesis, drug discovery, artificial intelligence, deep learning}



\maketitle


The development of a new drug is a complex process with high cost and risk.
It usually takes billions of dollars and ten or more years for an innovative drug to be developed and finally put on the market~\cite{wouters2020estimated}.
Innovation in synthetic chemistry plays an essential role in accelerating the process of drug discovery, providing the opportunity to gain more rapid access to biologically active, complex molecular structures in a cost-effective manner that can change the practice of medicine~\cite{blakemore2018organic,campos2019importance}.
During the first half of the twentieth century, most synthetic routes were developed by selecting appropriate starting materials after a trial-and-error search for suitable reactions to transform the chosen starting materials to the desired product.  In most cases, this approach was strongly dependent on the assumed starting point and did not allow an efficient search for the desired synthetic route~\cite{corey1991multistep}.
To design more reliable and efficient synthesis routes, retrosynthesis~\cite{corey1988robert,corey1991multistep}, which starts from the desired product and converts it to simpler precursor structures without defining the starting materials beforehand, has been an important technique for solving the planning of organic synthesis. 


\begin{figure*}[ht]
    \centering
    \includegraphics[width=0.98\textwidth]{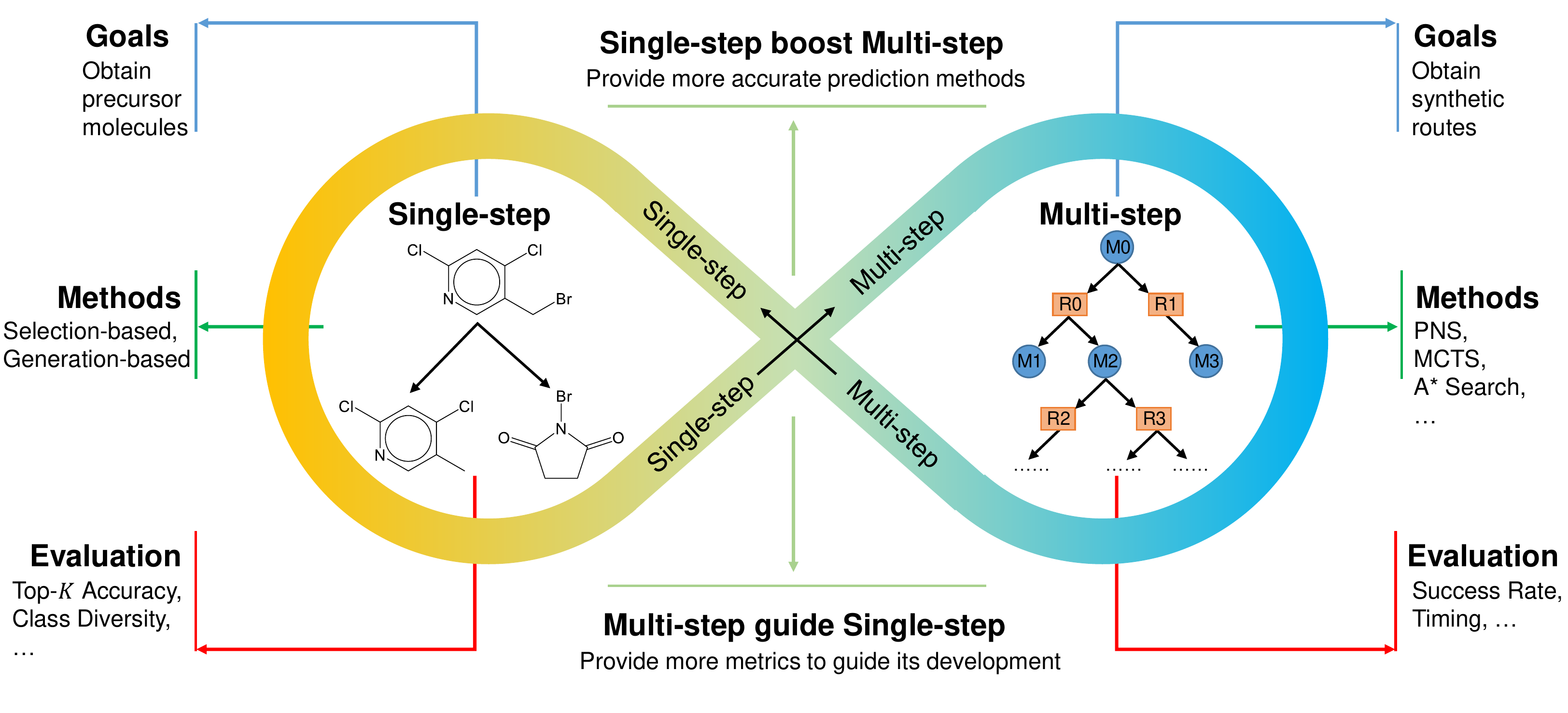}
    \caption{\textbf{The overview of single-step and multi-step retrosynthesis driven by artificial intelligence.}
    The retrosynthesis reaction of 5-(Bromomethyl)-2,4-dichloropyridine is shown in the single-step retrosynthesis, and an incomplete AND/OR tree is shown in the multi-step retrosynthesis.
    `M' nodes are OR nodes and represent molecules. 
    `R' nodes are AND nodes and represent the candidate reactions.
    }
    \label{fig:overview}
\end{figure*}

\begin{tcolorbox}[float*=t, floatplacement=t, width=\textwidth]
\subsection*{Box 1: Glossary of selected terms}

\textbf{Atom Mapping, Reaction Template, and Synthon.} The atom mapping numbers of the reactant and product represent the assignment of the atom before and after the reaction. By utilizing this information, researchers can extract the reaction template and synthon.
The reaction template is the substructures of molecules that actually participate in the reaction, and the synthon is the reactant substructure obtained by breaking the reaction center bond of the product.

\textbf{ECFP.} Extended-Connectivity Fingerprint (ECFP)~\cite{rogers2010extended} is a novel class of topological molecular fingerprints~\cite{todeschini2008handbook} and developed explicitly for activity modeling. It is a bit vector that contains indexed elements encoding physicochemical or structural properties, which allows it to be widely adopted for different analysis tasks, such as similarity searching, clustering, and virtual screening.

\textbf{SMILES.} Following the depth-first traversal over the molecular graph, a molecule can be represented by a string of ASCII characters called simplified molecular input line entry system~(SMILES). It can contain~(but not require) some information of stereochemistry, such as isomeric specification, configurations around tetrahedral centers, and many other types of chiral centers~\cite{david2020molecular}. A molecule can have multiple valid SMILES representations called ``randomized SMILES''.

\textbf{Molecular Graph.} Molecules can be intuitively represented by graph structures, where atoms are represented by nodes and bonds by edges. Mathematically, a molecular graph can be formally defined as $G=(A,F,E)$, where $|F| = N$, $|E| = M$, and $A \in \mathbb{R}^{N\times N}$ is the adjacency matrix. 
Researchers usually build node and edge feature matrices $F \in \mathbb{R}^{N\times D}$ and $E \in  \mathbb{R}^{M\times K}$ according to their needs, where $D$ and $K$ are the dimensions of node and edge features.

\textbf{Deep Neural Network.} A deep neural network is an artificial neural network built by multiple computational units called layers between the input and the output. By stacking simple but many transformations, it can be extremely powerful and widely adopted for different tasks nowadays.

\textbf{Graph Neural Network.} Graph neural networks~(GNN) are deep-learning methods that operate on the graph domain and can perform a series of graph analyses such as node classification, graph classification, and link prediction.
They usually consist of multiple GNN layers that follow the following computational units: propagation, sampling, and pooling modules.
These modules determine the mechanics of message passing on the graph.

\textbf{Transformer.} 
Transformer~\cite{vaswani2017attention} is the most popular machine translation model so far.
It is an end-to-end model following a stepwise and autoregressive encoder-decoder fashion. The key idea of the Transformer is the attention mechanism, which allows each token to capture the global information, which makes it quite suitable for SMILES representation.
 
\textbf{Data Augmentation.} Data augmentation, usually achieved by artificially representing the data in multiple forms, is a useful technique to increase the data volume, which can enrich the diversity of the data and build a more general model.

\textbf{AND/OR tree.} A search tree for multi-step retrosynthesis.
The root node of the tree is the target molecule, and the leaf nodes are usually commercially available starting materials.
An OR node represents a molecule, and an AND node represents a set of precursor molecules.

\end{tcolorbox}


In the early days of retrosynthesis, expert synthetic chemists could design synthesis routes with their familiar reactions.
However, owing to the enormous scale of chemical space, it is quite challenging to design a synthetic route for an unfamiliar molecule with a purely manual search. 
A recent study~\cite{szymkuc2016computer} estimates that more than 10, 000 transformations can be considered at each synthesis step, demonstrating how complicated the decision-making process is.
With the development of computer-aided technology since the 1970s, scientists started to design computer systems to integrate chemical reaction knowledge, including LHASA~\cite{pensak1977lhasa}, SYNLMA~\cite{johnson1989designing}, WODCA~\cite{gasteiger2000computer}, and Synthia~\cite{szymkuc2016computer}, to name a few.
These systems usually manually hard-coded reaction rules of specific areas to build a Network of Organic Chemistry~(NOC)~\cite{feng2018computational}, which largely hinders their wide application in real-world scenarios.
Recently, inspired by the success of data-driven artificial intelligence in various areas such as GO playing~\cite{chen2016evolution}, medical image classification~\cite{cai2020review}, and graph generation~\cite{guo2020systematic}, researchers began to solve retrosynthesis with deep learning approaches~\cite{baskin2016renaissance,chen2018rise} that automatically solve the single-step and multi-step planning both.
Heifets~\textit{et al.}~\cite{heifets2012construction} first introduced the proof-number search, a search algorithm of artificial intelligence for organic synthesis planning, showing the feasibility of applying an automatic planning technique to solve the multi-step retrosynthesis.
After that, Segler~\textit{et al.}~\cite{segler2017neural} first studied how to apply deep neural networks~\cite{yi2016study} to select an appropriate chemical rule for the product, which achieved a top-10 accuracy of 95\% in single-step retrosynthesis on a validation set of almost 1 million reactions.
Coupling this single-step retrosynthesis method with Monte Carlo tree search~(MCTS)~\cite{browne2012mcts}, they developed a fully automated multi-step retrosynthesis technique called 3N-MCTS~\cite{segler2018planning} that performed better and faster than any previous method.
These impressive advances all demonstrate the great potential of artificial intelligence in this field.
However, since the single-step solvers of these methods are either replaced with domain experts or specific to reaction rules, they cannot escape the limitations of high labor cost and poor generalization, which leads to the search for deep learning methods that do not rely on any expert experience or predefined reaction rule.

\begin{table*}[htbp]
  \centering
  \caption{A taxonomy of single-step retrosynthesis methods.}
  \resizebox{\textwidth}{!}{
\begin{tabular}{rrlll}
\toprule
\multicolumn{1}{l}{\textbf{Category}} & \multicolumn{1}{l}{\textbf{Goal}} & \textbf{Molecule Representation} & \textbf{Architecture/ Algorithm} & \textbf{Reported Works} \\
\midrule
\multicolumn{1}{l}{Reactant Selection} & \multicolumn{1}{l}{Select appropriate reactants} & SMILES & Bayesian inference + Transformer & \cite{guo2020bayesian} \\
      &       & Molecular Graph  & GNN & \cite{lee2021retcl}\\
\midrule
\multicolumn{1}{l}{Template Selection} & \multicolumn{1}{l}{Select appropriate reaction template} & Molecular Fingerprint & Highway network & \cite{segler2017neural,fortunato2020data} \\
      &       &    Molecular Fingerprint    & Molecular similarity & \cite{coley2017computer}\\
      &       &     Molecular Fingerprint   & Hopfiled Netwrok & \cite{seidl2022improving}\\
      &       & Molecular Graph  & GNN & \cite{ishida2019prediction,dai2019retrosynthesis,chen2021deep}\\
\midrule
\multicolumn{1}{l}{Semi-Template Generation} & \multicolumn{1}{l}{Break the product to synthons, and } & SMILES  & Transformer & \cite{wang2021retroprime}\\
      & \multicolumn{1}{l}{then generate reactants based on } & Molecular Graph  & GNN & \cite{shi2020graph,somnath2021learning}\\
      & \multicolumn{1}{l}{synthons} & Molecular Graph + SMILES & GNN + Transformer & \cite{yan2020retroxpert}\\
\midrule
\multicolumn{1}{l}{Template-Free Generation} & \multicolumn{1}{l}{Generate reactants using an} & SMILES  & LSTM & \cite{liu2017retrosynthetic}\\
      & \multicolumn{1}{l}{end-to-end approach} &   SMILES & Transformer & \cite{karpov2019transformer,zheng2019predicting,chen2019learning,yang2019molecular,lin2020automatic,tetko2020state,seo2021gta,kim2021valid,irwin2022chemformer,zhong2022root}\\
      &       & Molecular Graph  & GNN & \cite{sacha2021molecule}\\
      &       & Molecular Graph + SMILES & Transformer & \cite{mao2021molecular} \\
      &       & Customized Strings  & Transformer & \cite{mann2021retrosynthesis,ucak2021substructure,ucak2022retrosynthetic}\\
\bottomrule
\end{tabular}%
    }
  \label{tab:single_taxonomy}%
\end{table*}%


In recent years, there have been tremendous advances in AI-driven retrosynthesis techniques, both single-step and multi-step. In Fig.~\ref{fig:overview}, we provide an overview of these techniques, including their goals, methods, and evaluation metrics.
Although the work on single-step and multi-step is often published separately from each other, we believe that single-step and multi-step retrosynthesis are closely intertwined. If the single-step solvers could provide more accurate predictions, the success rate and search time of the multi-step methods would naturally decrease accordingly. In the meantime, since single-step ones ultimately serve the multi-step, researchers of multi-step retrosynthesis could propose new metrics for the single-step to guide its development. Therefore, these two technologies should be closely integrated and inseparable. In this review, we will focus on these existing AI-based retrosynthesis methods, aiming to provide a comprehensive overview of current advances and point out some possible future research directions. 

The following sections are organized as follows: for single-step and multi-step retrosynthesis both, we first list their goal and provide a thorough taxonomy of existing methods. Subsequent sections will analyze these methods in terms of their mechanism and performance and then introduce popular evaluation metrics for them, in which we also provide a detailed comparison among representative methods on several public datasets. In the next part we introduce popular databases and established platforms for retrosynthesis. Finally, this review concludes with a discussion about promising research directions in this field. A glossary of selected terms is provided in Box 1.

\section{Single-step Retrosynthesis Methods}
Generally speaking, it is ideal for single-step retrosynthesis to  directly obtain the complete structure of reactants by inputting a product molecule, but this is not necessarily the goal of single-step retrosynthesis.
In fact, existing single-step methods can be divided into two categories depending on whether to predict the complete reactant structure or not, namely the selection-based and generation-based.
1) Selection-based. Similar to the previous rule-based methods, by predefining reactant or reaction template candidates to be selected, these methods formulate retrosynthesis as a selection problem.
Since they utilize chemical knowledge to varying degrees, they are more likely to make valid and successful predictions on reactions that are similar to ones in the training set. However, it is impossible for them to make predictions outside the candidates, which largely limits their generalization ability.
2) Generation-based. Taking no chemical knowledge as priors, these methods usually generate the target reactants directly in the representation of string sequences or molecular graphs, which allows them to generalize to a wider variety of reactions. Based on the goal of single-step retrosynthesis, our proposed taxonomy of deep single-step retrosynthesis methods is shown in Table.~\ref{tab:single_taxonomy}. In this section, we aim to introduce the main idea of each category and specific methods.

\subsection{Selection-based Methods}
There are two types of selection-based retrosynthesis methods, reactant selection and template selection, which utilize chemical knowledge in varying degrees. Their ultimate goal is to select the appropriate reactant or reaction template instead of directly generating the reactant, which can be formulated as follows:
\begin{equation}
        {{\{r_i\}}_{i=1}^{n}}^* =  \mathop{\arg\max}\limits_{{\{r_i\}}_{i=1}^{n}}(f({\{r_i\}}_{i=1}^{n},p))
\end{equation}
and 
\begin{equation}
            t^* = \mathop{\arg\max}\limits_{t}(f(t,p))
\end{equation}
where $p$ is the desired product, ${\{r_i\}}_{i=1}^{n}$ is the set of reactant molecules, and $t$ is the reaction template.

\subsubsection{Reactant Selection}
These retrosynthesis methods directly select appropriate reactants from the molecule candidates by calculating the matching score between the product and the possible reactant.
There are some obvious advantages over other categories:
1) The candidates can be restricted to the commercially available molecules, which allows chemists to conduct practical experiments efficiently;
2) The validity of the molecules of the reactants is guaranteed;
3) It can be easily adapted to a specific set of molecular candidates.

Although there are currently few such methods, they have already achieved comparable results against others.
Since forward reaction prediction can be regarded as the inverse task of retrosynthesis prediction, Guo \textit{et al.}~\cite{guo2020bayesian} decided to use the forward prediction results to perform the retrosynthesis prediction.
They first proposed a workflow based on the inversion of the forward model into the backward one via Bayes's law of conditional probability. Specifically, the prediction result of retrosynthesis is obtained by the probability of getting the desired product through forward reaction prediction.
The reliability of the prediction relies on the high accuracy of the forward prediction~(top-1 accuracy of 90.4\% for Molecular Transformer~\cite{schwaller2019molecular}).
In the inference stage, they selected all reactant molecules in the dataset as the candidate set.

Similarly, Lee \textit{et al.}~\cite{lee2021retcl} also utilized forward prediction to aid retrosynthesis. 
However, they believed that the method should not rely exclusively on forward predictions but rather integrate both forward and backward predictions.
They introduced a framework called RetCL that first searches the reactant set candidates by a specified score function, and then ranks these possible reactant sets by a combined score of both forward and backward predictions.


Albeit these two methods, especially RetCL~\cite{lee2021retcl}, have achieved excellent results, such methods are severely limited by the candidate set.
The authors took all the reactants in the dataset~(including the test set) as candidates, implying that the correct answer must be in the candidate set when evaluating performance. This ideal situation rarely exists in real-world scenarios.
In order to make a fair comparison with other methods, its candidate set should be restricted to include only the molecules in the training set when measuring performance, which would make it much less effective. We will further elaborate on it in the Section Evaluation Methodology.


\begin{figure*}[ht]
    \centering
    \includegraphics[width=0.98\textwidth]{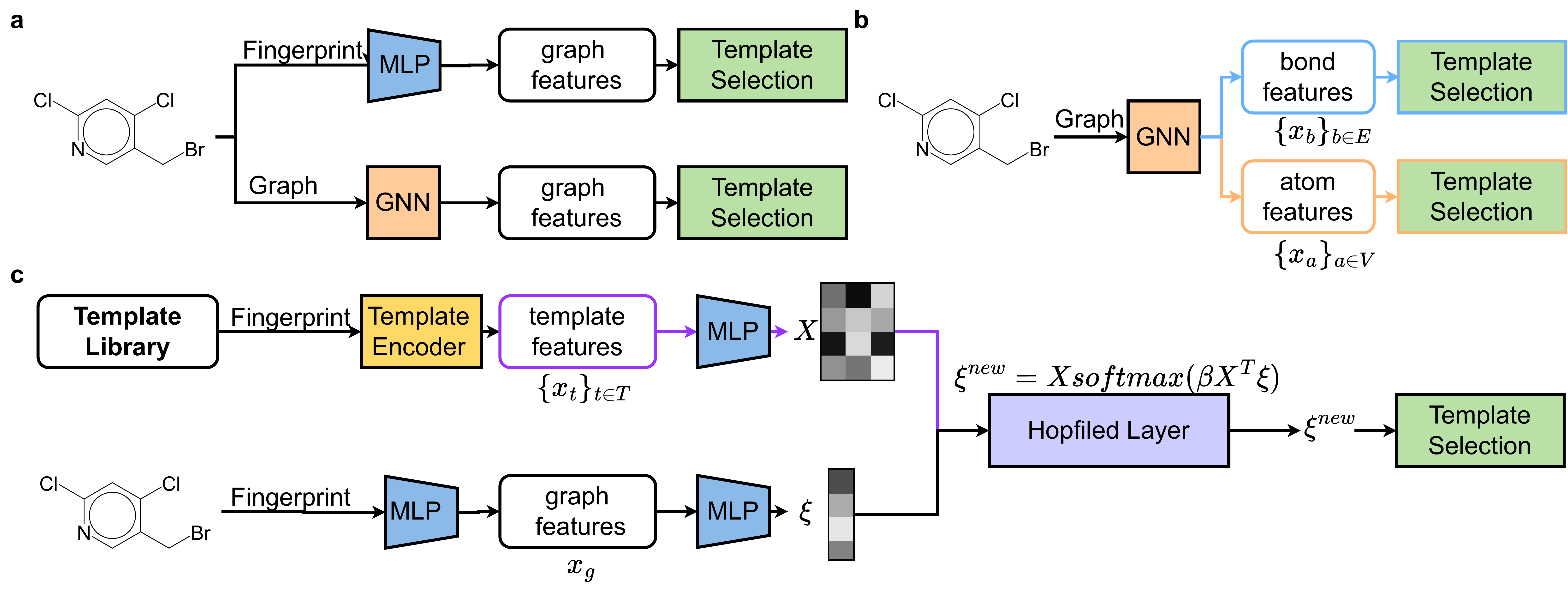}
    \caption{\textbf{The general workflow of template-based methods.}
    (a) Selecting the appropriate reaction template with extracted graph features.
    (b) Selecting the appropriate local reaction template with extracted node or bond features.
    (c) Modeling the retrosynthesis as a content-based retrieval problem.
    }
    \label{fig:template-based}
\end{figure*}


\subsubsection{Template Selection}
As described in Box 1, the reaction template is a subgraph pattern that describes the changes of atoms and bonds between a product molecule and its corresponding reactant(s).
Template-based methods formulate retrosynthesis as a classification problem of selecting a proper reaction template for the target product, which can be thought of as an interpolation of known reactions to novel substrates rather than an extrapolation to novel chemistry knowledge~\cite{coley2017computer}.
Although it is quite similar to the methods based on the selection of reactants, selecting the reaction template has the following advantages over selecting reactants:
1) Only one reaction template needs to be selected instead of selecting multiple reactants.
2) When only the reactions in the training set are considered as the candidate set, the coverage of reaction templates is much higher than that of reactants.
Because of the appearance of RDChiral~\cite{coley2019rdchiral} that can extract reaction templates automatically, there is no need to manually hard-encoding reaction templates by experts compared with the traditional reaction rules. In Fig.~\ref{fig:template-based}, we display the general workflow of existing template-based methods.

Neuralsym~\cite{segler2017neural} is not only the first template-based, but also the first attempt to apply neural networks to retrosynthesis. Although it was proposed as a rule-based method initially, it has been proven to work as an effective template-based method by Dai~\textit{et al.}~\cite{dai2019retrosynthesis}.
As illustrated in Fig.~\ref{fig:template-based}a, taking ECFP molecular fingerprint~\cite{rogers2010extended} as the input and highway network~\cite{srivastava2015highway} as the deep model, it could calculate an applicable probability for all reaction templates in the candidate set, which showed better performance than the previous expert systems and great potential of AI-driven retrosynthesis. This method has been widely adopted as the single-step solver of most recent multi-step retrosynthesis methods.

Since there are some templates with few examples in the dataset, Fortunato~\textit{et al.}~\cite{fortunato2020data} proposed a pretrain and data augmentation strategy for reaction templates to expand the chemistry scope covered by the training set. 
The strategy was performed by matching reaction templates extracted from the training set with products one by one to obtain an augmented training set with a many-to-many relationship between products and reaction templates. Using the same model framework as Neuralsym, a model was first pretrained on the multi-label task of matching the input molecule to multiple optional reaction templates on the augmented dataset, and then finetuned on the classification task of selecting the appropriate reaction template on the original dataset. 
This data augmentation strategy effectively improved the performance of rare templates, but still failed to deal with reaction templates that were missing in the training set. 

To improve the performance of few-show and zero-shot reaction template prediction, Seidl~\textit{et al.}~\cite{seidl2022improving} developed a new method that did not consider templates as distinct categories, but could leverage structural information about template.
They first proposed the concept of recasting the classification problem of reaction templates into a content-based retrieval problem.
Most existing template-based methods calculated the template matching scores by the following equation:
\begin{equation}
    \pmb{\hat{y}} = softmax(\pmb{W}\pmb{h}^m(\pmb{m}))
    \label{eq:non-cross-tb}
\end{equation}
where $\pmb{h}^m(\pmb{m}))$ was a neural network called \textit{molecule encoder} that mapped a molecule representation to a vector of size $d_m$. $\pmb{W} \in \mathbb{R}^{K \times d_m}$ was the last layer of the molecule encoder and $K$ was the number of reaction templates. Seidl~\textit{et al.} believed that this approach ignored the similarity among reaction templates and prevented generalization over them.
Therefore, as shown in Fig.~\ref{fig:template-based}c, they proposed the \textit{template encoder} to map each template to a vector of size $d_t = d_m$, and converted Eq.~(\ref{eq:non-cross-tb}) to the following equation:
\begin{equation}
    \pmb{\hat{y}} = softmax(\pmb{h}^t(\pmb{T})\pmb{h}^m(\pmb{m}))
    \label{eq:cross-tb}
\end{equation}
where $\pmb{h}^t(\pmb{T}))$ was a neural network called \textit{template encoder} that mapped the template library to a matrix of size $K\times d_t$.
To solve such a retrieval problem, they adopted a fingerprint-based modern Hopfield network~(\textit{MHN})~\cite{ramsauer2020hopfield,widrich2020modern} to associate relevant templates to product molecules. Experiments conducted by them showed that their method had a considerable advantage over others when there were no or few examples in the training set, which proved that learning the similarity among templates was important for predicting rare templates. The authors also pointed out that this framework could be applied to other methods based on graph representation or SMILES.

There are two major drawbacks for Nerualsym and its derivations:
1) Neuralsym is a completely full-connected neural network that lacks the interpretability of the prediction reason.
2) Because of the lack of complete chemical structural information, molecular fingerprints are usually considered inadequate as input for deep learning.
Aiming to address the above two issues,
Ishida~\textit{et al.}~\cite{ishida2019prediction} proposed a new framework based on graph convolutional networks~(GCN)~\cite{kipf2016semi} with an architecture-free visualization method called integrated gradients~(IG)~\cite{sundararajan2017axiomatic} for highlighting reaction-related atoms.
By representing molecules as heterogeneous graphs with atom features and introducing a more advanced neural network, it not only achieved better performances, but also succeeded in finding some atoms related to the reaction with IG. 
They defined IG $\pmb{I}$ as follows:
\begin{equation}
    \pmb{I}(x_t)=\frac{\Delta x_t}{M}\sum_{k=1}^M\nabla_x S_l(\frac{k}{M}(\Delta x_t+x^0)
    \label{eq:IG}
\end{equation}
where $x_t$ is an atom of a molecule, $x^0$ is a reference atom whose feature matrix is a zero matrix, $M$ is the number of a division of the reference atoms, and $S_l(\frac{k}{M}(\Delta x_t+x^0)$ is a score value of the softmax layer's $l$-th neuron for the predicted reaction template.

The concurrent work of Dai~\textit{et al.}~\cite{dai2019retrosynthesis} also adopted graph representation and introduced interpretability. They proposed Conditional Graph Logic Network~(GLN), a conditional graphical model defined with logic rules, where the logic variables were chemistry knowledge about reaction templates. It was designed to solve the joint probability of the reaction template and precursors. By calculating the inner product of each atom embedding in the target molecule with the matched template subgraph embedding, they visualized the prediction of probabilistic modeling of the reaction center to interpret their models.

Instead of relying on neural networks, Coley \textit{et al.}~\cite{coley2017computer} formulated the reaction template selection as a similarity comparison problem.
Based on the molecular similarity calculated by molecular fingerprints, they first search for the most similar product to the test molecule in the training set. By applying the reaction template of that product, the possible reactant can be easily obtained. Finally, the order of reaction templates was determined by a combined score of reactant and product similarities.

The work mentioned above is all based on the global features of the input molecules to make predictions.
However, for the reason that the changes during the chemical reaction occur mostly locally,
Chen \textit{et al.}~\cite{chen2021deep} pointed out that the use of global features may yield undesired focus on the details not directly related to the reaction. 
As shown in Fig.~\ref{fig:template-based}b, they designed a new template-based workflow by locally deriving the reaction templates and evaluating the suitability of these local templates at all enumerated possible reaction centers for a molecule.
As the goal of the workflow was to predict a set of local templates at each possible reaction center, the desired model should focus more on local information.
Therefore, Chen~\textit{et al.} developed a model called \textit{LocalRetro}, which is mainly composed of message passing neural network~(MPNN)~\cite{gilmer2017neural} and global reactivity attention~(GRA).
MPNN was a spatial convolutional framework that allowed atoms and bonds to learn their local environment.
GRA was a variety of multi-head self-attention applied in Transformer~\cite{vaswani2017attention} to capture the information from the remote atoms and bonds.
LocalRetro, which made good use of the nature of chemical reactions, achieved the best performance among all existing template-based methods.

\subsection{Generation-based Methods}
Generation-based retrosynthesis methods rely on no chemical knowledge, where only a single product is taken as the input, and reactants are usually represented by heterogeneous molecular graphs or SMILES strings.
According to whether the generation is done in one step or not, they can be divided into template-free methods and semi-template methods.


\begin{figure*}[htbp]
    \centering
    \includegraphics[width=0.98\textwidth]{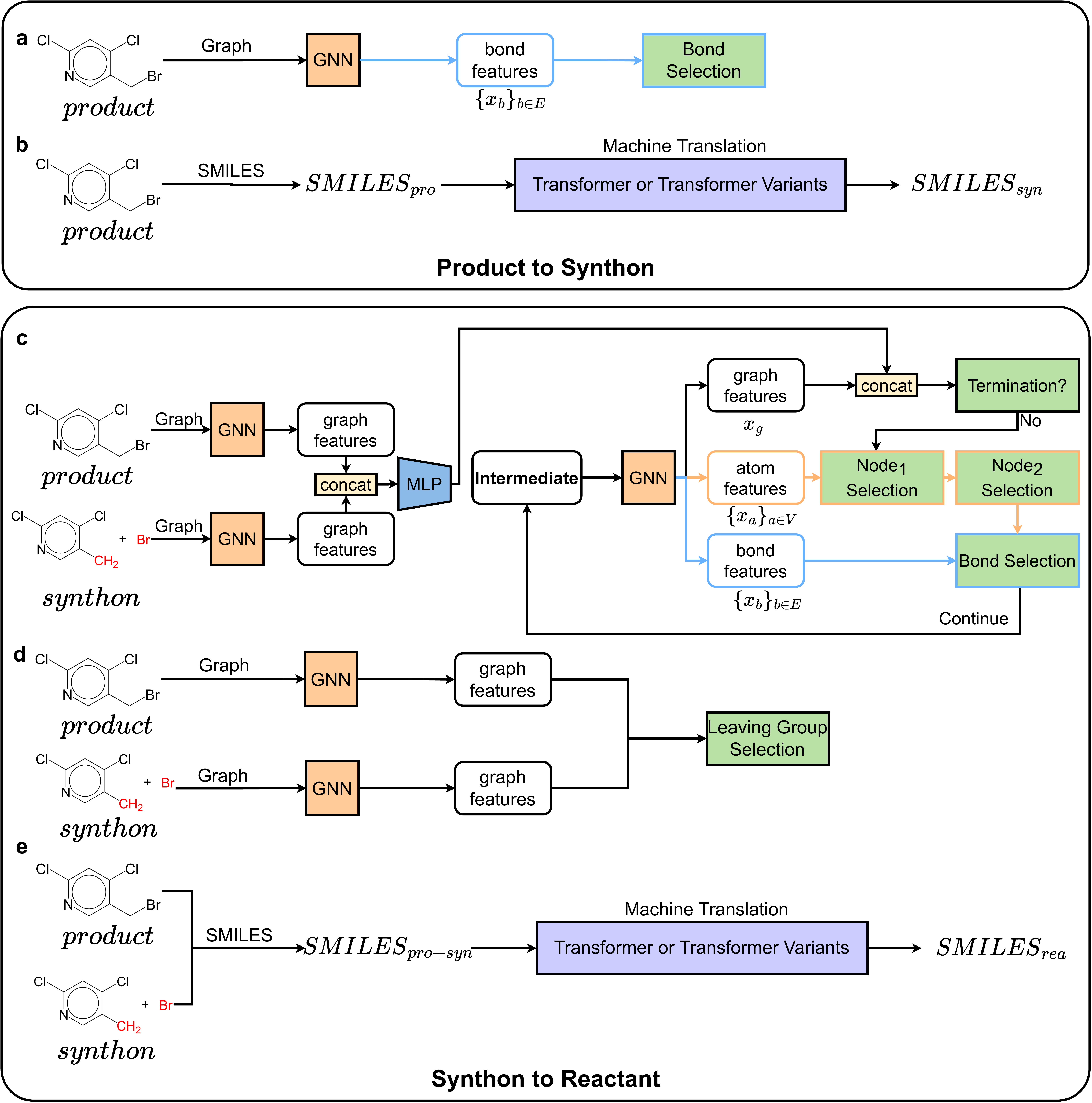}
    \caption{\textbf{The general workflow of semi-template generation methods.}
    (a) Modeling the product-to-synthon stage as a link prediction problem.
    (b) Modeling the product-to-synthon stage as a machine translation problem.
    (c) Modeling the synthon-to-reactant stage as a graph generation problem. The model iteratively determines if the generation should terminate. If not, the model would select appropriate nodes and connect them with a appropriate bond.
    (d) Modeling the synthon-to-reactant stage as a leaving group selection problem.
    (e) Modeling the synthon-to-reactant stage as a machine translation problem.
    Theoretically the free combination between product-to-synthon methods and synthon-to-reactant methods is possible.
    }
    
    \label{fig:semi_template}
\end{figure*}

\subsubsection{Semi-Template Generation}
Since the chemical reaction in the datasets is atom-mapped, the transformations of atoms and bonds during the reaction can be automatically identified by comparing the product to their corresponding reactants. The retrosynthesis can be resolved by predicting these transformations. Following this idea, semi-template methods divide the retrosynthesis into two steps: first, identify the reaction center to get intermediate molecules called synthons, and then complete synthons to reactants.
Here for simplicity, the product is abbreviated as P, the synthon as S, and the reactant as R.These two-step transformations are denoted as P2S and S2R.
When the P2S stage succeeds, due to the correctly recognized reaction center and the one-to-one mapping between synthons and reactants, the accuracy can be much higher than other methods, which makes the accuracy of the P2S stage much more important.
As the focus of the P2S stage is to identify the reaction center, it is usually formulated as a link prediction~\cite{daud2020applications,rossi2021knowledge} task, while the S2R stage varies.
The S2R stage is aimed at completing the synthons, which can be solved by a lot of different approaches, such as directly generation, attaching leaving groups, machine translation, and so on. The general workflow of semi-template methods is shown the Fig.~\ref{fig:semi_template}.
Theoretically, the P2S and S2R approaches can be freely combined.



Shi~\textit{et al.}~\cite{shi2020graph} proposed the first semi-template generation method that is completely based on graph representation.
First, they broke the target product into several synthons and then attached new atoms to the synthons to accomplish the retrosynthesis~(Fig.~\ref{fig:semi_template}a).
To obtain the atom and bond features, they applied the Relational Graph Convolutional Network (RGCN)~\cite{schlichtkrull2018modeling} to it. 
As for the graph generation in the second step, Shi~\textit{et al.} leveraged the assumption of Markov Decision Process~(MDP)~\cite{puterman1990markov}, which satisfies the Markov property that $p(S^i\vert S^{i-1},z) = p(S^i\vert S^{i-1},...,S^0,z)$. 
The MDP formulation means that each action is only conditioned on the graph that has been modified so far~(Fig.~\ref{fig:semi_template}c), which implies its generation process is autoregressive.

Somnath~\textit{et al.}~\cite{somnath2021learning} combined the ideas of semi-template generation and reactant selection, proposing the GraphRetro, whose first step is to identify the reaction center~(Fig.~\ref{fig:semi_template}a), while the second step is to pick a correct leaving group to attach to the synthon~(Fig.~\ref{fig:semi_template}d). They defined the leaving groups as the substructures that can help synthons expand into valid reactants by attaching them. Although this approach cleverly circumvents the problem of generating in the semi-template generation, it also inherits the disadvantages of both the semi-template and selection-based methods, which will be further elaborated in the Evaluation Methodology Section.

Yan~\textit{et al.}~\cite{yan2020retroxpert} adopted molecular graphs and SMILES in the two steps, respectively.
First they proposed a variant of GAT~\cite{velickovic2017graph}, called EGAT, to integrate the features of chemical bonds in molecule graphs to obtain better graph representation~(Fig.~\ref{fig:semi_template}a). 
Since the previous practice of judging only one most probable reaction center was only applicable to reactions with a single or two reactants,  they introduced an additional task to predict the number of reaction centers, which enabled them successfully handled the reactions with three or more reactants.
To be specific, they first predicted how many reaction centers would be needed.
Based on this prediction result, select the chemical bonds that should be broken according to the probability. 
In the S2R stage, they followed the sequence generation formula by treating it as a machine translation task~(Fig.~\ref{fig:semi_template}e).

Wang~\textit{et al.}~\cite{wang2021retroprime} further adopted the machine translation for both steps of semi-template generation~(Fig.~\ref{fig:semi_template}b,e). They introduced several useful approaches to improve the performance without changing the model architecture of Transformer~\cite{vaswani2017attention}, which can be concluded as three points: (1) Since only the reaction centers need to be identified in the P2S stage, which means the translation from scratch is not required, they manually tagged the atoms of the reaction centers and identified the reaction centers by predicting these tags. (2) Based on the reaction center, they also added extra label information of whether it is a reaction center and whether it is connected to the reaction center to the input SMILES of the S2R stage, which can be utilized to align the input and output SMILES. (3) The previous method only selected the most probable synthon as the input of the S2R stage, which inevitably made the overall accuracy lower than the top-1 accuracy of the P2S stage. Aiming to address it, the authors first proposed the ``Mix and Match'' strategy, selecting multiple predicted synthons as the input of the S2R stage to increase the accuracy during the inference.

\begin{figure*}[htbp]
    \centering
    \includegraphics[width=0.98\textwidth]{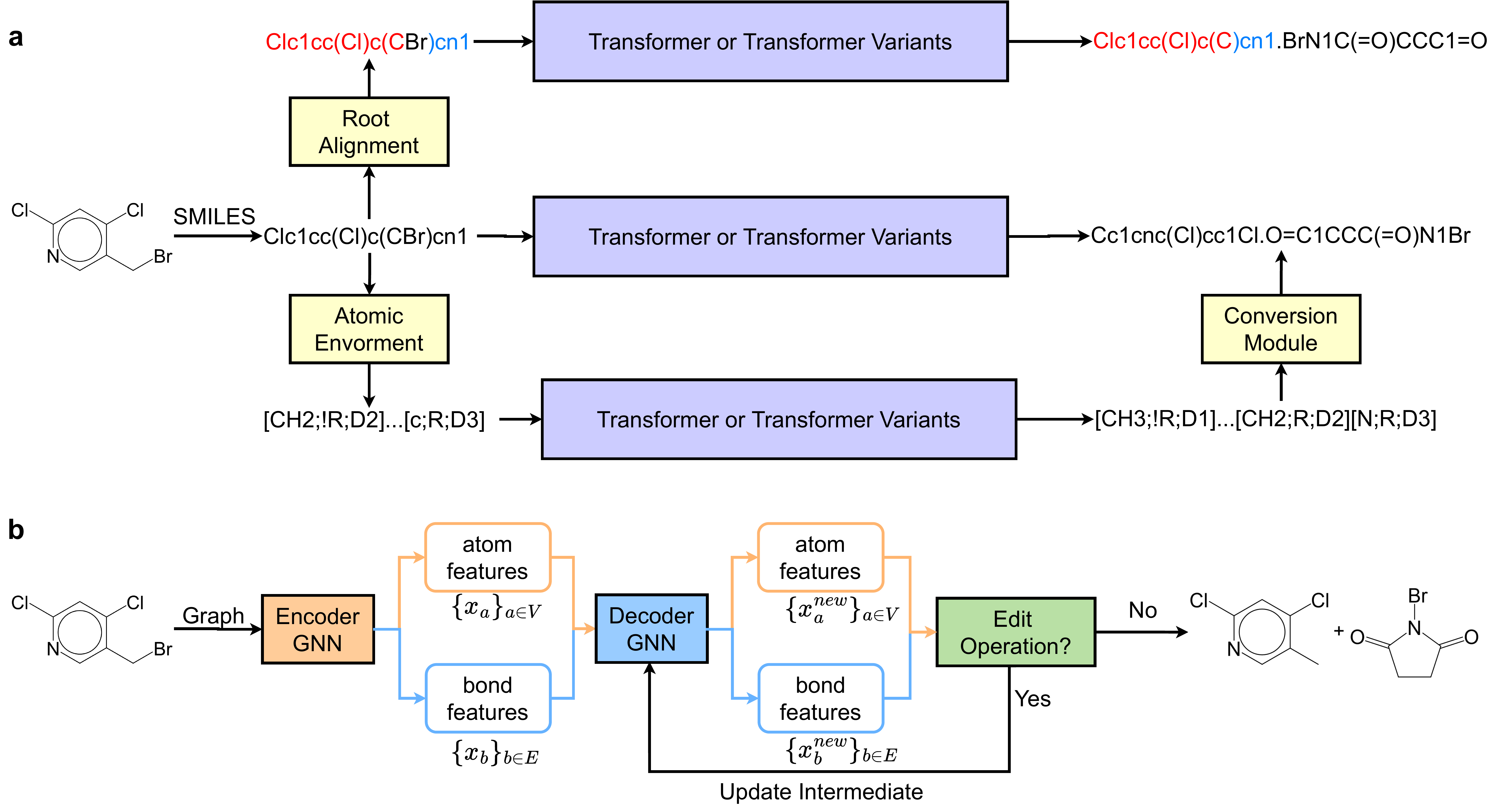}
    \caption{\textbf{The general workflow and difference of template-free generation methods.}
    (a) Modeling the retrosynthesis as a fully machine translation problem. There are majorly three string representations for molecules: canonical SMILES, atomic environment fingerprint representation, root-aligned SMILES.
    (b) Modeling the retrosynthesis as a fully graph edit problem. This is quite similar to the graph generation problem in the synthon-to-reactant stage. The model iteratively selects the graph action. If it is a editing action, update the intermediates and select next graph action, until the termination action is selected.
    }
    \label{fig:template_free}
\end{figure*}


\subsubsection{Template-Free Generation}
Template-free methods, \textit{i.e.}, fully end-to-end methods, formulate retrosynthesis as a sequence generation problem. 
The general workflow of template-free methods is shown the Fig.~\ref{fig:template_free}.
Depending on the representation of molecules, the sequence can be a series of SMILES tokens or molecular edit actions on the molecule graph. 
By adopting SMILES representation, a popular paradigm of existing retrosynthesis methods is to formulate retrosynthesis prediction as a sequence-to-sequence translation problem~\cite{yang2020survey}.
This translation process is usually token-by-token and autoregressive, \textit{i.e.}, based on the input product SMILES and the already decoded reactant SMILES, to obtain the next possible reactant token.
Researchers first create a token dictionary with the molecules in the training set.
Based on this dictionary and neural networks, they can get a probability distribution of the next possible token by inputting the product SMILES tokens and currently decoded reactant SMILES tokens. 
In order to improve the performance of retrosynthesis, researchers usually make improvements in the following perspectives: applying more robust models, using more effective training strategies, or improving the input and output of the models. 

By formulating retrosynthesis as a machine translation problem, Liu~\textit{et al.}~\cite{liu2017retrosynthetic} proposed the first template-free retrosynthesis model.
They designed an Encoder-Decoder architecture based on long short-term memory (LSTM)~\cite{van2020review} cells. 
Inspired by the attention mechanism~\cite{parikh2016decomposable,kim2017structured}, they calculated the attention based on the tokens of the product and treated it as an input of the decoder, achieving better performance than the rule-based expert system. 
But this model architecture was soon replaced by Transformer that is entirely driven by the attention mechanism.
Inspired by successful forward reaction prediction with Transformer~\cite{schwaller2019molecular}, Karpov~\textit{et al.}~\cite{karpov2019transformer} introduced the Transformer to the retrosynthesis and achieved better results than previous LSTM-based methods, which strongly demonstrated the strength of Transformer~(Fig.~\ref{fig:template_free}a).

As retrosynthesis is modeled as a translation problem, the output sequence may be grammatically invalid sentences, \textit{i.e.} prone to be invalid output molecules.
In fact, 12.2\% of top-1 predictions and 22.0\% of top-10 predictions by the vanilla Transformer are invalid~\cite{zheng2019predicting}.
To deal with it, Zheng \textit{et al.}~\cite{zheng2019predicting} proposed a self-corrected retrosynthesis predictor~(SCROP) that is a combination of two Transformers.
The first one served as the retrosynthesis predictor like the previous method, and the other as the SMILES syntax corrector. 
When training the second model, they constructed a training library that consisted of a set of input-output pairs, where the inputs were invalid SMILES predicted by the former retrosynthesis predictor, and the outputs were the ground-truth reactants.
While this approach largely reduced the invalid rate of predictions, it did not provide a significant accuracy improvement.

As mentioned above, forward reaction prediction can be treated as an inverse task of retrosynthesis prediction. Unlike Guo~\textit{et al.}~\cite{guo2020bayesian} that relied entirely on forward predictions, Wang~\textit{et al.}~\cite{wang2020forward} proposed the concept of ``forward verification'', that is, using the high-accuracy forward reaction prediction to verify the retrosynthesis prediction.
They conducted experiments on multiple sets of combined models, all successfully improving the top-1 accuracy of retrosynthesis.
The concurrent work of Kim~\textit{et al.}~\cite{kim2021valid} employed a similar concept called ``Cycle Consistency'', but further incorporated it into the network architecture and training process. They developed a Transformer-based network to train the forward reaction prediction and retrosynthesis simultaneously. Since the vast majority of network parameters for both tasks were shared, the training process could be seen as adopting a unique data augmentation strategy.
In order to generate diverse reactant candidates, they additionally introduced a learnable multinomial latent variable $z \in {1,...,K}$.
During the inference stage, they scored both forward reaction and retrosynthesis by ordering the reactant candidates based on the following likelihood:
\begin{equation}
    y^* = \arg\max(p(z\vert x)p(y\vert z,x)p(\tilde{x}=x\vert z,y))
\end{equation}
Experiments showed that the proposed method further reduced invalid rates and improved the diversity of predictions compared with previous ones.

Since a molecule can have multiple valid SMILES representations, many researchers performed data augmentation to improve the performance of seq2seq models~\cite{bjerrum2017smiles}.
Among the existing works, Tetko~\textit{et al.}~\cite{tetko2020state} conducted the most detailed study on different data augmentation strategies.
They first classified data augmentation strategies based on SMILES enumeration into four categories:
1) Augmentation of products only, 
2) Augmentation of products and reactants/reagents,
3) Augmentation of products and reactants/reagents followed by shuffling of the order of reactant/reagents, 
4) On the basis of 3), mixing forward and reverse reactions. 
Their experiments demonstrated that the more complex the data augmentation strategy was applied, the better the generalization ability of the model.
In addition, they first proposed the data augmentation applied to the test set, \textit{i.e.}, adopting different SMILES strings of a molecule as the input and then ranking all the predictions uniformly.
In our research, we found that this approach can not only effectively improve the accuracy of model prediction but also significantly reduce the invalid rate, which will be further elaborated in Section Evaluation Methodology.

The pretrain-finetune strategy of Transformer for text generation has been proven to be extremely effective~\cite{devlin2018bert}, so some researchers tried to apply it to retrosynthesis as well. 
The pretrain tasks can be majorly divided into two types:
1) The input and the output represent the same or similar content, which helps the model master the syntax;
2) The output is modified to make the pretrain task closer to the downstream task.
For the first type, the existing representative pretrain methods that can be applied on SMILES are BERT~\cite{devlin2018bert} and X-MOL~\cite{xue2021x}, though they are not proposed for retrosynthesis. 
BERT proposed to mask a part of tokens in the input sentence and let the network predict the masked part, which could help the model understand the meaning of various types of tokens.
X-MOL selected another valid and equivalent SMILES representation of the input molecule as the output of the network, which was aimed to allow it to learn the complex grammar rules of SMILES.
Combining these two pretrain strategies, Irwin~\textit{et al.}~\cite{irwin2022chemformer} succeeded in largely improving top-1 accuracy. However, this approach also significantly reduced the top-10 accuracy, indicating that it simultaneously reduced the diversity of model predictions.
For the second type of pretraining, Chen~\textit{et al.}~\cite{chen2019learning} proposed two different strategies to generate outputs: random pretraining and template-based pretraining. The former only involved randomly breaking a bond of the input molecule to generate the output SMILES, while the latter involved breaking a bond and attaching a new group according to a matched reaction template, both of which yielded similar improvements.

The above studies all illustrate that Transformer performed well on the retrosynthesis task. However, some researchers questioned the direct application of SMILES to the vanilla Transformer, arguing that this approach did not fully utilize both the characteristics of the molecule.
By adding an extra graph encoder, Mao~\textit{et al.}~\cite{mao2021molecular} proposed the representation fusion of SMILES sequences and molecular graphs to enrich the features.
Seo~\textit{et al.}~\cite{seo2021gta} went one step further and proposed Graph-Truncated Attention~(GTA) that could modify the attention mechanism based on molecular graph topology. More specifically, by modifying the attention masks, they forced different attention heads in self-attention to pay attention to atom pairs at a different distance, and tried to make the cross-attention mask as close as possible to the atom-mapping mask.

\begin{table*}[htbp]
  \centering
  \caption{A taxonomy of multi-step retrosynthesis methods.}
  \resizebox{\textwidth}{!}{
    \begin{tabular}{rlll}
    \toprule
    \multicolumn{1}{l}{\textbf{Search Algorithm}} & \textbf{Method} & \textbf{Highlights} & \textbf{Single-Step Solver} \\
    \midrule
    \multicolumn{1}{l}{Proof-Number Search} & PNS~\cite{heifets2012construction}   & The first application of artificial intelligence to multi-step retrosynthesis & Expert \\
          & DFPN-E~\cite{kishimoto2019depth} & Balance the number of AND and OR nodes & Expert \\
    \midrule
    \multicolumn{1}{l}{Monte Carlo Tree Search} & 3N-MCTS~\cite{segler2018planning} & Evaluate the pathway quality via double-blind AB tests & Neuralsym \\
          & AutoSyn~\cite{lin2020automatic} & Combine MCTS with the template-free model Transformer & Transformer \\
          & ReTReK~\cite{ishida2022ai} & Introduce retrosynthesis knowledge to guide the search direction & GCN \\
    \midrule
    \multicolumn{1}{l}{A* Search} & Retro*~\cite{chen2020retro} & Consider both the informed cost and estimated future value of nodes & Neuralsym \\
          & RetroGraph~\cite{xie2022retrograph} & Represent the search process as a directed graph rather than a search tree & Neuralsym \\
    \midrule
    \multicolumn{1}{l}{Others} & Self-Improved~\cite{kim2021self}  & A end-to-end framework for improve single-step solver & Neuralsym \\
          & Hyper-graph~\cite{schwaller2020predicting} & Apply forward model to filter suggestions  & Transformer \\
          & SimulatedExp~\cite{schreck2019learning} & Apply reinforcement learning to find a optimal search policy & Neuralsym \\
    \bottomrule
    \end{tabular}%
    
    }
  \label{tab:multi_taxonomy}%
\end{table*}%

Another group of researchers has worked on improving SMILES, arguing that there are more suitable inputs or outputs for Transformer than the original SMILES.
As a molecule has multiple valid SMILES representations, Sumner~\textit{et al.}~\cite{sumner2020levenshtein} selected the input-output pairs with the smallest edit distance as the training data, effectively aiding the cross-attention mechanism in the Transformer. 
Mann~\textit{et al.}~\cite{mann2021retrosynthesis} and Ucak~\textit{et al.}~\cite{ucak2022retrosynthetic} defined the new hand-crafted representations of the molecule that contain more features and are more suitable for machine translation~(Fig.~\ref{fig:template_free}a). 
However, a huge drawback of using hand-crafted representations is that there is no guarantee that the hand-crafted representations can be converted back to SMILES.
Ucak~\textit{et al.} converted their representations to SMILES by additionally training a translation model or by similarity matching, which not only did not guarantee the success of the translation, but also added additional time consumption. 

In contrast to the manually designed representations, Zhong~\textit{et al.} decided to improve SMILES into a more suitable representation for reaction prediction.
There are two common problems in previous SMILES-based retrosynthesis methods: the inability to guarantee a one-to-one mapping between input and output, especially when applying data augmentation, and the large discrepancy between input and output SMILES. To address these two issues, Zhong~\textit{et al.}~\cite{zhong2022root} proposed root-aligned SMILES~(R-SMILES), which ensured a strict one-to-one mapping and small edit distance between input and output SMILES by aligning the root atoms of them.
As shown in Fig.~\ref{fig:template_free}a,  the colored tokens in the input and output SMILES represent their common parts. Combining R-SMILES with the data augmentation strategy proposed by Tetko~\textit{et al.}~\cite{tetko2020state}, they achieved the best performance among template-free methods to date.
This approach is theoretically applicable to any SMILES-based retrosynthesis method.

Unlike the above SMILES-based template-free methods, Sacha \textit{et al.}~\cite{sacha2021molecule} proposed a novel framework using molecular graph editing. As shown in Fig.~\ref{fig:template_free}b, starting from the product molecule, they performed a sequence of editing operations, including \textit{EditAtom, EditBond, AddAtom, AddBenzene, Stop}. Compared with SMILES-based methods,  the graph structure provides richer information for the model. Moreover, the graph editing operations avoids generating reactants from scratch, which allows  the model to produce less invalid predictions and explore chemical space efficiently.

\section{Multi-step Retrosynthesis Methods}
While single-step methods are continuously being improved, most molecules in the real world cannot be synthesized within one step. The average industrial pharmaceutical synthesis route contains 8.1 steps~\cite{carey2006analysis}, and some complex academic targets or natural products may require over 100 steps~\cite{nicolaou1996classics}. Such a colossal space poses challenges for efficient searching and planning of multi-step retrosynthesis.
In multi-step retrosynthesis, the algorithm usually needs to build a search tree or a directed acyclic graph starting from the target molecule and ending to the commercially available building blocks. Based on the optional reactions provided by single-step retrosynthesis methods, the search is guided by the designed objective function and repeated until the raw materials are all commercially available. Considering the exponentially growing search space as the route grows, it poses a considerable challenge for researchers to find a solution within the limited route length.
As the number of possible synthesis routes is often astronomical, it is desirable to identify the route that minimizes some user-specified objective functions, such as the synthetic cost and the length of the reaction route. 
Depending on the adopted search algorithm, our proposed taxonomy of multi-step retrosynthesis methods is shown in Table~\ref{tab:multi_taxonomy}.
This section aims to provide a detailed review of these methods.

\subsection{Proof-Number Search}
Unlike blind traversals like depth-first traversal and breadth-first traversal, the heuristic algorithm relies on the heuristic function and the current state to decide the next action.
With the estimated distance between the current node and the target calculated by the heuristic function, the algorithm can select the most urgent node to analyze and expand the search tree.
Heifets~\textit{et al.}~\cite{heifets2012construction} are the first to model multi-step retrosynthesis planning as a discrete state-space search problem, showing it is amenable to heuristic search techniques.
By adopting the proof-number search~(PNS)~\cite{allis1994searching}, an AND/OR tree search algorithm, they modeled it as a two-player zero-sum game where one must synthesize target molecules as possible, and the other tries to prevent it. In practice, the first player picks the reaction rule to synthesize the target molecule first, and the other player selects one of the precursor molecules generated by the reaction rule. Two players alternatively play moves until the end of the game. The first player wins if and only if he can construct all precursor molecules, implying a valid synthetic route.
As illustrated in Box 1, an AND/OR consists of AND nodes representing reactions and OR nodes representing molecules. In an AND/OR tree, a node is called \textit{proven} if it is guaranteed to win for the first player. In other words, if the molecule is a available building block, an OR node is proven. Similarly, a node is called \textit{disproven} if the first player cannot synthesize it. 

Kishimoto~\textit{et al.}~\cite{kishimoto2019depth} pointed out a significant search space imbalance in PNS applied to multi-step retrosynthesis prediction. Since a molecule can be synthesized by various reactions, while most reactions contain only one precursor molecule, the branching factor of OR nodes is much larger than the one of AND nodes, resulting in a lopsided search space. In the lopsided search space, PNS have difficulties in identifying moves with higher chances of leading to proofs. To address this phenomenon, they propose a depth-first proof-number search with Heuristic Edge Initialization~(DFPN-E), which assigns a heuristic cost to an edge from an OR node to an And node. However, both of these PNS methods use human-designed heuristics and do not yet exploit the full potential of artificial intelligence.


\subsection{Monte Carlo Tree Search}
Based on the proposed Neuralsym~\cite{segler2017neural}, Segler~\textit{et al.}~\cite{segler2018planning} combined three different neural networks together with Monte Carlo Tree Search~(MCTS)~\cite{browne2012mcts} to perform multi-step retrosynthesis prediction.
MCTS is a heuristic search algorithm that repeats the selection, expansion, rollout, and update steps.
(1) Selection. Starting from the root of the search tree, the algorithm selects the most urgent node on the basis of the current position values. In the context of MCTS, each node represents a set of precursor molecules.
(2) Expansion. The authors adopted two neural networks to guide the search in promising reaction rules as well as to verify whether these reaction rules are feasible or not.
(3) Rollout. If the molecules in the current position are not building blocks, authors will use the third neural network to quickly find a solution to synthesize them within the maximal depth of the search tree.
(4) Update. Depending on the difficulty of the molecule synthesis in the previous step, the algorithm receives a reward and updates the position values in the tree.
Compared with PNS, MCTS focuses more on balance between exploration and exploitation~\cite{kishimoto2019depth}. 
Afterwards, Lin~\textit{et al.}~\cite{lin2020automatic} also combined MCTS with the template-free method Transformer and successfully reproduced several synthetic routes in the literature.

Ishida~\textit{et al.}~\cite{ishida2022ai} proposed a promising research direction to integrate various portions of retrosynthesis knowledge into multi-step retrosynthesis planning. On the basis of MCTS architecture proposed by Segler~\textit{et al.}~\cite{segler2018planning}, they made two major improvements: 1) replaced the highway network with a more advanced GCN-based policy network to make better single-step predictions; 2) defined a series of retrosynthesis scores to evaluate promising search directions, including a convergent disconnection score~(CDScore), an available substances score~(ASScore), a ring disconnection score~(RDScore), and a selective transformation score~(STScore). The introduction of these scores led to a search preference for convergent reactions, reactions with available reactants, ring construction reactions, and reactions with few by-products, respectively. The domain knowledge does not necessarily improve the success rate, but it can guide the algorithm in the direction that the chemists want to search.

\begin{figure*}[htbp]
    \centering
    \includegraphics[width=\textwidth]{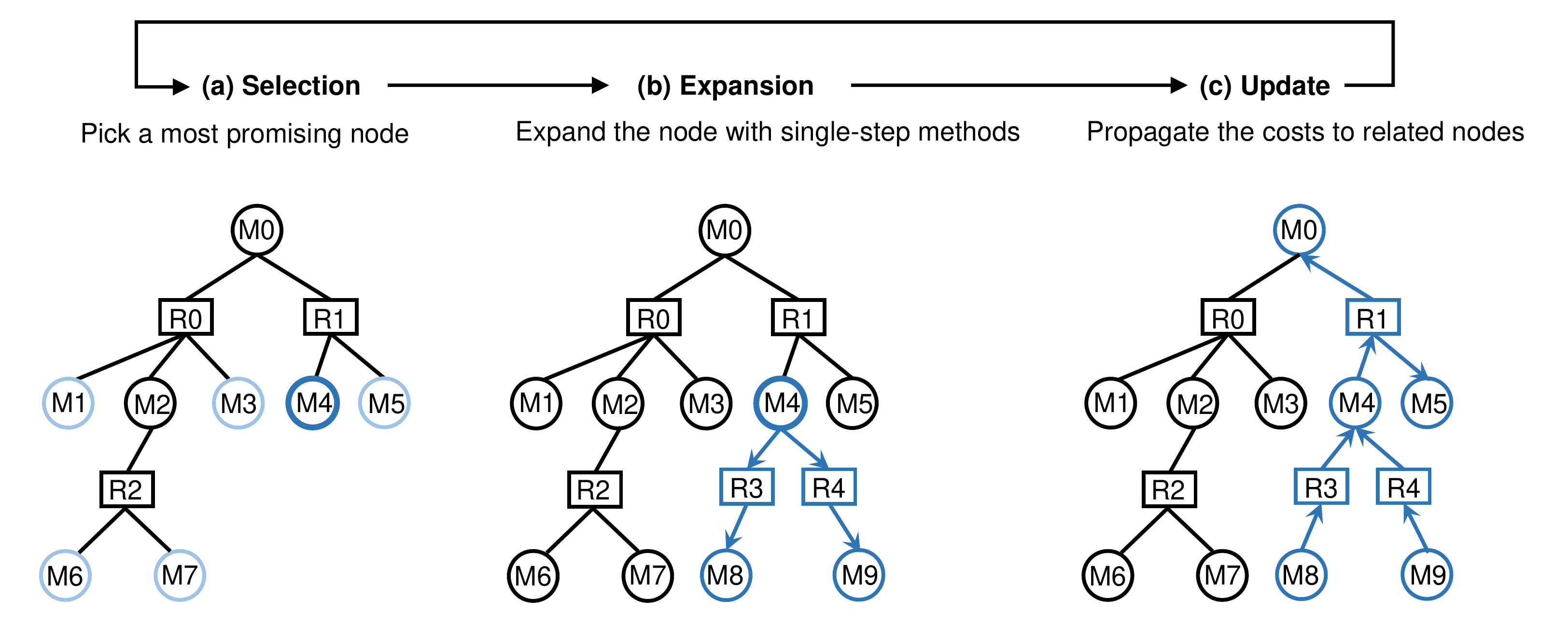}
    \caption{\textbf{The workflow of A* Search with an AND/OR tree.}
    The `M' nodes are OR nodes and represent molecules.
    The `R' nodes are AND nodes and represent candidate reactions.
    In the selection step (a), the leaf nodes that are not building blocks are colored blue and the selected node is bold.
    }
    \label{fig:A_star_search}
\end{figure*}

\subsection{A* Search}
As previous methods depend on online value estimation and are not efficient enough, Chen~\textit{et al.}~\cite{chen2020retro} proposed the Retro* algorithm, which combined A*~\cite{hart1968formal} search with the AND/OR tree and models the retrosynthesis problem as a single-player game for focusing on the global value estimation. 
Compared with the general heuristic algorithm that only estimates the cost to the target, A* search additionally considers the informed cost of the current position from the starting position, which enables it to find a better solution.
As shown in Fig.~\ref{fig:A_star_search}, A* search repeats the selection, expansion, and update steps to find a solution.
The whole process is very similar to MCTS, but with the rollout step reduced.
This is because that the authors directly predicted the future value of the nodes by an additional neural network during the selection process, which is much more efficient than the iterative rollout strategy in MCTS.
Therefore, the search algorithm can find the desired expansion direction more efficiently and generalize to the unknown reaction data. 

For the previously mentioned tree search algorithm, there is a significant flaw: there could be multiple duplicated molecule nodes in a search tree, which significantly require more iterations and limit the success. Therefore, Xie~\textit{et al.}~\cite{xie2022retrograph} proposed a graph-based search method RetroGraph, which is combined with a novel GNN-guided policy to eliminate intra-target duplication and improve the success rate. By representing the search process as a directed graph, different reaction nodes can point to the same molecule nodes, i.e., share the same precursor molecules, dramatically reducing the number of iterations and the number of nodes. 






\subsection{Others}
Existing methods tend to optimize single- or multi-step solvers individually, without considering both as a whole. Therefore, Kim~\textit{et al.}~\cite{kim2021self} proposed a universal end-to-end framework that aims to improve the single-step solver for maximizing the success rate of multi-step search algorithms.  On top of the original single-step solver and multi-step search algorithm, they introduced two additional models: a reference backward model to discard unrealistic predictions and a forward reaction model to perform data augmentation. To be specific, After gathering a collection of reaction $C$ from the original single-step solver and multi-step search algorithm, the reference model would first discard unrealistic suggestions in $C$,  and then the forward model would generate new products by inputting the reactants in $C$ to get a new collection of reaction $C'$. The single-step solver would be trained with $C \cup C'$ to make better predictions. These steps could be repeated until the model achieves optimal results. The authors conducted experiments on the Retro*~\cite{chen2020retro}, where although the accuracy of single-step model did not differ much, the success rate of reaction pathway showed a significant improvement. 

To choose the most promising direction of the retrosynthesis, Schwaller~\textit{et al.}~\cite{schwaller2020predicting} proposed an additional metric, round-trip accuracy, to evaluate single-step methods. It refers to the ratio of how many of the predicted products generated by another forward reaction model is the desired product, which verifies the validity of the reaction. 
By coupling it with the SCScore~\cite{coley2018scscore} that quantifies the molecular synthetic complexity, they scored the suggestions provided by the single-step solver and chose the most promising suggestion to update the search graph.

Although the vast majority of current multi-step search methods are based on heuristic search, non-heuristic algorithms are still directions worth exploring, one of which is deep reinforcement learning~\cite{li2017deep}.
Reinforcement learning~(RL) is about an agent interacting with the environment by trial and error to learn an optimal policy for sequential decision-making problems, which has made a great impact in chess, shogi, and GO playing~\cite{silver2017mastering,mnih2015human,silver2018general}.
Following this fashion, Schreck~\textit{et al.}~\cite{schreck2019learning} formulated the retrosynthesis as a single-player game. Starting from a random policy, the model is asked to estimate the accurate cost for each molecule and find the better solution by repeating the game many times, resulting in a constantly improved policy. This learned policy through simulated experience can enable the search to explore in more meaningful directions.

\section{Evaluation Methodology}
Here we aims to provide comprehensive quantitative analyses of current representative single-step and multi-step retrosynthesis algorithms on publicly available datasets.

\subsection{Single-step Retrosynthesis}

In order to rigorously determine whether one model is better or worse than another, it is necessary to develop benchmarking metrics that can be evaluated for models trained on publicly available datasets. 
In this section, we experimentally make a detailed comparison among different retrosynthesis methods quantitatively and then introduce other evaluation metrics, including round-trip accuracy, class diversity, and invalid rate.

To maximize the fairness of the comparison, we followed the following principles in our experiments: 1) Use the code provided by the authors for the experiments whenever possible. If the authors did not provide the code, we would use the other faithful reimplementation.
2) In our experiments, we used the default parameters suggested by the authors, including the number of network layers, learning rate, and other hyperparameters. Although some methods can get higher performance after a careful selection of hyperparameters, we believe that the robustness of the method is also important.

\begin{figure}[htbp]
    \centering
    \includegraphics[width=0.48\textwidth]{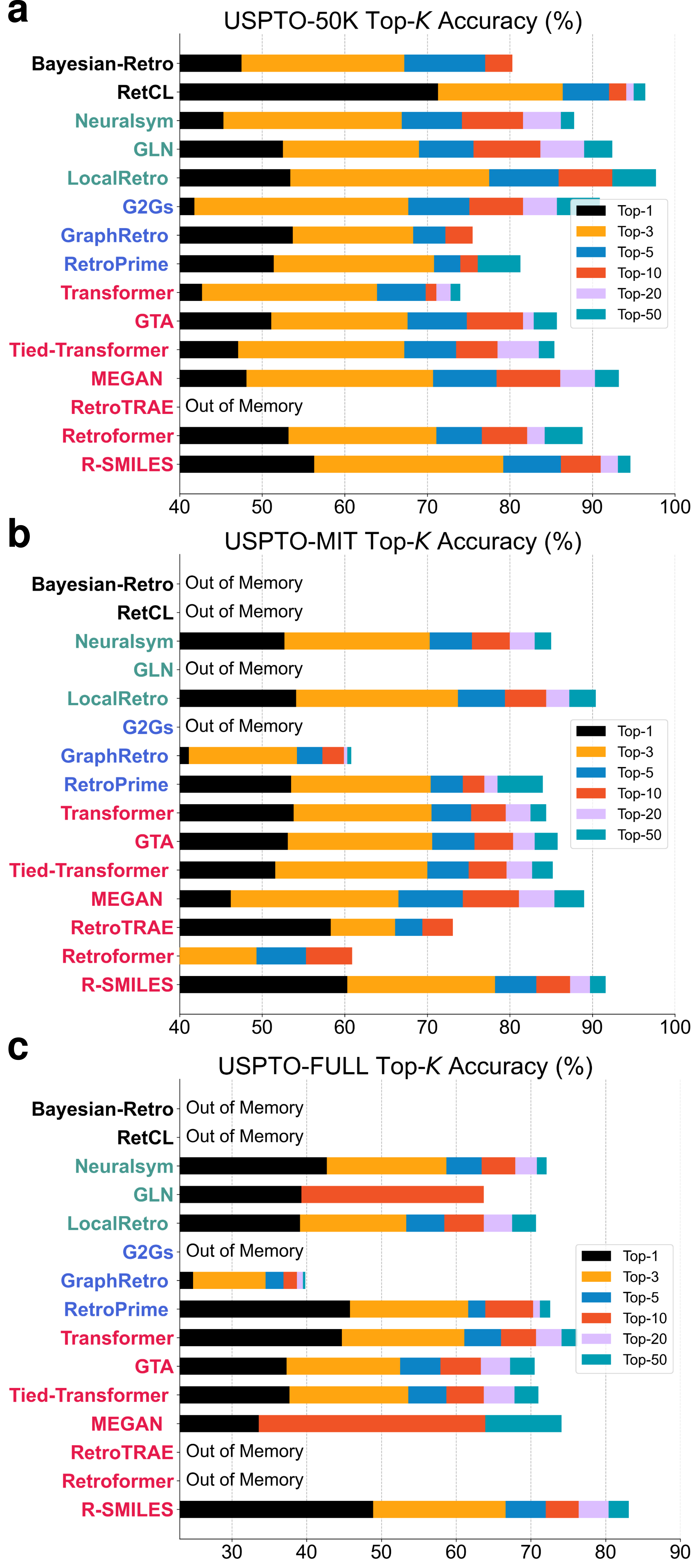}
    \caption{\textbf{Top-$K$ single-step retrosynthesis results on USPTO-50K (a), USPTO-MIT (b), and USPTO-FULL (c) datasets.}
    The black, orange, blue, red, purple, green color bars represent top-1, top-3, top-5, top-10, top-20, top-50 accuracy, respectively.
    The black methods are reactant selection ones,  the green are template selection ones, the blue are semi-template generation ones, and the red are template-free generation ones.
    ``Out of Memory'' means that the methods are irreproducible because they require too much computer memory.
    }
    \label{fig:topk_acc}
\end{figure}

\subsubsection{Datasets}
There are three popular public benchmark datasets for single-step retrosynthesis, USPTO-50K~\cite{schneider2016s},  USPTO-MIT~\cite{jin2017predicting} and  USPTO-FULL~\cite{dai2019retrosynthesis}, which are all derived from the United States Patent and Trademark Office~(USPTO)-granted patents database~\cite{lowe2012extraction}.
USPTO-50K is a high-quality dataset containing about 50,000 reactions with accurate atom mappings between products and reactants. 
USPTO-MIT contains about 400, 000 reactions as the training set, 30, 000 reactions as the valid set and 40, 000 reactions as the test set. Although it has much more data, the diversity of reactions is not as high as that of USPTO-50K. For example, no chiral molecules are present in this dataset. 
USPTO-FULL is a much larger dataset for chemical reactions, consisting of about 1,000,000 reactions. However, as the atom mapping is obtained by the Indigo toolkit, some atom mappings are wrong, which directly leads to incorrect reaction templates and synthons.


\subsubsection{Quantitative Evaluation}
The top-$K$ accuracy is the most valuable and widely used among all the currently available metrics. It means the percentage of the correct answer appearing in the top-$K$ predictions.
This metric is not only intuitive for researchers but also usually the training task for deep models. 
It is intuitive to think that the higher the top-1 accuracy, the better method, but sometimes top-10 and top-50 accuracies carry more weight, since in multi-step retrosynthesis it is valuable to provide multiple options for each step to find the global optimal solution.
In Fig.~\ref{fig:topk_acc}, we list the top-$K$ accuracy of the representative methods on different public datasets. 
Considering that the reaction type is usually unavailable in realistic scenarios, the performance is all without the reaction type information.
Most of these results are obtained from their published papers, while some are obtained from additional experiments we conducted. The implementation and specific accuracy of each method are available online at https://github.com/otori-bird/DeepRetrosynthesis.

First, we do not dwell on specific methods but focus on comparisons among different categories, drawing the following conclusions: (1) In the case of a small amount of data, namely the USPTO-50K dataset~(Fig.~\ref{fig:topk_acc}a), selection-based methods generally outperform generation-based methods. This is because the chemical priors they introduce successfully guide the model's reaction predictions, i.e., limit the model's search space, while generation-based methods need to find an answer in a larger search space, which may even be an invalid molecule. (2) In the case of a huge amount of data, that is, on the USPTO-MIT and USPTO-FULL datasets~(Fig.~\ref{fig:topk_acc}b, c), we can see the effect of the selection-based method is gradually equaled or even surpassed by the generation-based method. We believe that they are constrained by the introduced chemical knowledge in turn now. Taking template-based methods as an example, for the USPTO-FULL dataset, the reaction templates in the training set can only cover about 77\% of the reactions in the test set, which means that the remaining part can never be predicted correctly. Similarly, if we limit the available reactants to the molecules in the training set, the coverage of reactant-selection methods on the test set of USPTO-50K, USPTO-MIT, and USPTO-FULL datasets would be 17.9\%, 65.2\%, and 72.8\%, respectively.
However, generative-based methods do not suffer from these limitations.  (3) As the semi-template generation methods divide the retrosynthesis into two separate steps, which irreversibly makes their overall accuracy limited by the accuracy of the first step, their top-10 and top-50 accuracies are worst among different categories.

Now we are making some comparisons among specific methods. There are some interesting findings: (1) Although GLN~\cite{dai2019retrosynthesis} and LocalRetro~\cite{chen2021deep} outperform Neuralsym~\cite{segler2017neural} obviously on the USPTO-50K, what surprises us greatly is that among template-based methods,  these leads are shrinking or even reversing as the amount of the data increases, which may imply that the final performance of template-based methods is less related to the representation of molecules or the model architecture when data is sufficient. 
This can be interpreted in two ways:
i.) These complex network architectures accelerate the convergence but also limit the search space.
ii.) The pre-extracted chemical information in the molecular fingerprint is more helpful for the model to pick the suitable template when dealing with a huge amount of data. 
(2) The similar phenomenon can also be found in comparing template-free methods. Although GTA~\cite{seo2021gta} and Tied-Transformer~\cite{kim2021valid} are variants of the Transformer, they only perform significantly better than Transformer on the USPTO-50K. Only R-SMILES~\cite{zhong2022root} keeps outperforming Transformer, which is granted as it applies data augmentation and root alignment to the Transformer without changing the model architecture.
Since the performance on large datasets is more convincing and of greater interest, it is uncertain to state that Neuralsym is worse than the later template-based methods.
(3) Compared with another semi-template method RetroPrime~\cite{wang2021retroprime}, GraphRetro~\cite{somnath2021learning} performs poorly in the two large datasets. As mentioned above, GraphRetro~\cite{somnath2021learning} is a combination of semi-template and selection-based, which makes it suffer from the limitation of both the accuracy of the P2S stage and the coverage of the leaving groups.

\subsubsection{Other Evaluation Metrics}

\begin{itemize}
    
     \item Round-trip Accuracy: The round-trip accuracy quantifies what percentage of the retrosynthetic suggestions are considered valid by the forward model. 
     This indicator was formally proposed by a multi-step retrosynthesis study~\cite{schwaller2020predicting}, and its concept is now widely adopted in multiple single-step methods. 
     For example, Kim~\textit{et al.}'s Tied-Transformer and Lee~\textit{et al.}'s RetCL both take it as an auxiliary scoring mechanism for model predictions, i.e., their methods consider both backward and forward scores when evaluating model predictions.
     We see it as an indication of the multi-step research contributing to the development of single-step research.
     
    
    \item Class Diversity: Class diversity measures the diversity of the model's predictions when there is no given reaction type. 
    It is unrealistic for human experts to judge the reaction type for each prediction result, which naturally leads to the question of whether a neural network can be trained to judge it. By adopting the GNN-based or transformer-based reaction type classifier,  Chen~\textit{et al.}~\cite{chen2019learning}, Wang~\textit{et al.}~\cite{wang2021retroprime}, and Schwaller~\textit{et al.}~\cite{schwaller2020predicting} illustrated the diversity of their predictions, respectively.
    Among them, the discriminator proposed by Schwaller~\textit{et al.}~\cite{schwaller2021mapping} achieved an accuracy of 98.2\%.
    
    
    \item Top-$K$ invalid Rate: Top-$K$ invalid rate means the percentage of the chemically invalid prediction appearing in the top-$K$ predictions. Since selection-based methods are impossible to provide invalid predictions, this metric is mainly used for judging generation-based methods. Predicting invalid sentences is very common in machine translation methods. For example, the top-10 invalid rate was as high as 22\% for Liu~\textit{et al.}'s seq2seq, which indirectly also determines the upper bound of top-10 accuracy. Although SCROP~\cite{zheng2019predicting} largely reduced the invalid rate by adding a syntax correcting model, it did not provide significant accuracy improvement. This problem is now addressed by the data augmentation on the test set~\cite{tetko2020state}. In our experiments, it was almost impossible for the Transformer to provide an invalid output when the test set was applied to 20$\times$ data augmentation. The only problem with this approach is the additional computation.
    
    
\end{itemize}

\begin{table*}[htbp]
  \centering
  \caption{Performance comparison among the multi-step metthods in the \textit{USPTO} dataset. We compare each method at the success rate of different iteration limit. The numbers of iterations, reaction (Rec.) nodes, and molecule~(Mol.) nodes are showed in average under the limit of 500.~\cite{xie2022retrograph}}
  \resizebox{\textwidth}{!}{
    \begin{tabular}{ccccccccc}
    \toprule
    \multirow{2}[4]{*}{\textbf{Method}} & \multicolumn{5}{c}{\textbf{Success Rate of Iteration Limit (\%) ↑}} & \multirow{2}[4]{*}{\textbf{Iteration ↓}} & \multirow{2}[4]{*}{\textbf{Rec. Nodes ↓}} & \multirow{2}[4]{*}{\textbf{Mol. Nodes ↓}} \\
\cmidrule{2-6}          & 100   & 200   & 300   & 400   & 500   &       &       &  \\
    \midrule
    DFPN-E~\cite{kishimoto2019depth} & 50.53 & 58.42 & 64.21 & 68.42 & 75.26 & 208.12 & 3123.33 & 4635.08 \\
    MCTS~\cite{segler2018planning}  & 43.68 & 47.37 & 54.74 & 58.95 & 62.63 & 254.32 & -     & - \\
    Retro*~\cite{chen2020retro} & 52.11 & 66.32 & 76.84 & 81.05 & 86.84 & 166.72 & 2927.92 & 4174.52 \\
    RetroGraph~\cite{xie2022retrograph} & \textbf{88.42} & \textbf{97.89} & \textbf{98.95} & \textbf{99.47} & \textbf{99.47} & \textbf{45.13} & \textbf{674.22} & \textbf{500.43} \\
    \bottomrule
    \end{tabular}%
    }
  \label{tab:multi_uspto}%
\end{table*}%

\begin{table*}[htbp]
  \centering
  \caption{Performance comparison among the multi-step metthods in the \textit{USPTO-EXT} dataset. We compare each method at the success rate of different iteration limit. The numbers of iterations, reaction (Rec.) nodes, and molecule~(Mol.) nodes are showed in average under the limit of 100.~\cite{xie2022retrograph}}
  \resizebox{\textwidth}{!}{
    \begin{tabular}{cccccccccc}
    \toprule
    \multirow{2}[4]{*}{\textbf{Method}} & \multicolumn{6}{c}{\textbf{Success Rate of Iteration Limit (\%) ↑}} & \multirow{2}[4]{*}{\textbf{Iteration ↓}} & \multirow{2}[4]{*}{\textbf{Rec. Nodes ↓}} & \multirow{2}[4]{*}{\textbf{Mol. Nodes ↓}} \\
\cmidrule{2-7}          & 10    & 20    & 30    & 40    & 50    & 100   &       &       &  \\
    \midrule
    Retro*~\cite{chen2020retro} & 42.47 & 48.79 & 51.84 & 53.63 & 55.00  & 57.89 & 48.49 & 790.49 & 1136.51 \\
    RetroGraph~\cite{xie2022retrograph} & \textbf{50.84} & \textbf{58.05} & \textbf{62.05} & \textbf{64.26} & \textbf{66.89} & \textbf{72.89} & \textbf{37.25} & \textbf{491.97} & \textbf{373.80} \\
    \bottomrule
    \end{tabular}%
    }
  \label{tab:multi_uspto_ext}%
\end{table*}%

\subsection{Multi-step Retrosynthesis}
Due to the ambiguous definition of ``good synthesis route'', evaluating multi-step methods is much more complicated than single-step ones. 
Segler~\textit{et al.}~\cite{segler2018planning} have conducted a double-blind AB test where 45 graduate-level organic chemists have to choose a better route from those in the literature and those provided by their algorithm. Although very intuitive and practical, this approach is too time-consuming and expensive to be used for evaluation on large-scale datasets.
In this section, we will present the experimental details of the multi-step retrosynthesis, including the adopted datasets, evaluation metrics, and the quantitative evaluation of several representative methods as well as a public framework for benchmarking multi-step methods.



\subsubsection{Datasets}
The starting material in multi-step retrosynthesis is usually defined as the commercially available compound~\cite{genheden2020aizynthfinder}.
eMolecules~\footnote{https://www.emolecules.com/} that consists of 231 million available molecules, usually serves as the building block library for researchers.
In addition, ZINC~\cite{irwin2020zinc20} is also a reliable database of available materials, which contains about 1.3 billion purchasable molecules.

Heifets~\textit{et al.}~\cite{heifets2012construction} created the first public benchmark dataset that contains 20 synthesis routes derived from undergraduate organic examinations.
Considering the small scale of the test data, Segler~\textit{et al.}~\cite{segler2018planning} and Kishimoto~\textit{et al.}~\cite{kishimoto2019depth} selected 497 and 897 instances for evaluation, respectively.
Based on USPTO~\cite{lowe2012extraction} database and commercially available building blocks from eMolecules,
Chen~\textit{et al.}~\cite{chen2020retro} first created a large benchmark dataset called \textit{USPTO} that contains 299,902 training routes, 65,274 validation routes, and 189 test routes.
After that, Xie~\textit{et al.}~\cite{xie2022retrograph} built an extra test set called \textit{USPTO-EXT}, whose size is ten times the size of \textit{USPTO} test set.

\subsubsection{Quantitative Evaluation}
Some researchers only tested their proposed methods on a few instances to illustrate the validity of their methods, which is contingent and unconvincing.
We believe that a method should be evaluated on a public dataset in terms of success rate and time to perform the convincing evaluation.
Thanks to the work of Xie~\textit{et al.}~\cite{xie2022retrograph}, we can display and compare recent advanced multi-step retrosynthesis search algorithms on the same dataset in Table~\ref{tab:multi_uspto} and Table~\ref{tab:multi_uspto_ext}. These methods all adopted Neuralsym as their single-step solver. The number of iterations in these tables refers to the number of single-step solver calls, which can measure the computational cost as it occupies over 99\% of the running time~\cite{chen2020retro}.

First of all, we can find that RetroGraph~\cite{xie2022retrograph} is the only one of these methods that uses directed graphs instead of search trees, which allows it to eliminate redundant intermediate molecule nodes and speed up the algorithm.
Guided by the more advanced GNN, it also outperforms others at the success rate of any iteration limit. In addition, Retro*~\cite{chen2020retro} and RetroGraph~\cite{xie2022retrograph}, which utilized A* search algorithm, achieved the most comprehensive results. On the other hand, MCTS~\cite{segler2018planning} is not only the worst in terms of success rate, but also slowest due to the four-step update process.
However, considering that the route diversity, i.e., the ability of the method to discover new synthetic routes, is not measured in these experimental data, we cannot directly conclude that MCTS is the worst method.

\subsubsection{Public Benchmark Framework PaRoutes}
As shown in Table~\ref{tab:multi_taxonomy}, most of the multi-step search methods use Neuralsym~\cite{segler2017neural} as their single-step solver for fairness of comparison, but this approach does not allow the academic community to know the power of current state-of-the-art retrosynthesis. For example, combining the best methods in single-step and multi-step retrosynthesis might outperform any previous result.
Genheden~\textit{et al.} proposed a framework called PaRoutes~\cite{genheden2022paroutes} for benchmarking multi-step methods. It examines the performance in several aspects such as route quality, search speed, and route diversity. With implemented DFPN, MCTS, and Retro* algorithms in the open-source retrosynthesis software AiZynthFinder~\cite{genheden2020aizynthfinder}, they found that DFPN is significantly inferior to others and MCTS performs slightly better than Retro* in terms of route quality and route diversity. 
Since PaRoutes provides an open and fair evaluation platform for researchers, we encourage the later researchers to evaluate their methods with it.


\section{Database and Platform for Retrosynthesis}
In recent years, many mature and complete data databases have emerged. The vast amount of chemical information and the application of big data technology have provided great convenience for the researchers to develop a fully automatic computer-aided synthesis planning~(CASP) tool. Utilizing the retrosynthesis techniques mentioned above, several research groups have also developed their CASP platforms to assist chemists.
Some of these databases and platforms are open-source and publicly available, while others are supported by industrial companies and require commercial access.
In order to demonstrate the current progress of retrosynthesis,  this section introduce popular reaction databases as well as several well-established retrosynthesis platforms.



\subsection{Reaction Database}
\subsubsection{Public Database}
United States Patent and Trademark Office~(USPTO)-granted patents is the predominant current open-source reaction database in the field of machine learning, containing 1,939,253 reactions that were extracted by text-mining from U.S. patents published between 1976 and 2016~\cite{lowe2012extraction}.
However, with a large number of incorrect or duplicate reactions, the data quality is quite uneven. Furthermore, the atom mapping in reactions is automatically generated by the Indigo toolkit and is not guaranteed to be correct.
Researchers usually clean their data before using this database, deriving different datasets for single-step and multi-step retrosynthesis, which will be further elaborated on below.

Cheminformatics Elsevier Melbourne University lab~(ChEMU)~\cite{he2020extended,he2020overview} is a manually annotated database of organic reaction texts in 1500 patents.
In addition to reactants and products, the annotated text also contains the reagent catalyst, solvent, time, temperature, yield, and so on. 

\subsubsection{Commercial Database}
Operated by Elsevier, Reaxys~\cite{goodman2009computer} is a major commercial subscription reaction database. It enables users to search for synthetic routes of the target molecule as well as the corresponding reaction conditions like temperature and catalyst. It also supports the addition of a series of user-specified conditions to the search for reactions, such as specific substructures, yields, temperatures, etc. The information for each reaction includes the corresponding patent or literature, which allows users to better understand the reaction principle. The suppliers, prices, and shipping time of the raw materials of the synthesis route are also available.

Similar to Reaxys, Chemical Abstracts Service~(CAS) is a huge database of both organic and inorganic reactions, covering detailed information of over 145 million single-step and multi-step reactions since 1840. It also supports the reaction search with additional user-specified conditions like Reaxys.

Another subscription database is Pistachio. Similar to the USPTO database, it is also built based on a text-mining approach that extracts reaction information from patents. However, in addition to the larger data, it is maintained and kept updating by the NextMove Software. With over 9 million reactions extracted from the patents in United States and European Patent Office, it provides a friendly interface to query and analyze chemical reactions.


\subsection{Retrosynthesis Planning Platform}

\subsubsection{Public Platform}
With template-based single-step solver and the root-parallelized MCTS, Coley~\textit{et al.} developed a open-source software for CASP called ASKCOS~\cite{coley2019robotic}, which was trained with millions of reactions in both USPTO and Reaxys. Its full workflow consists of three main steps: 
1) based on the reaction templates extracted from USPTO and Reaxys, ASKCOS propose a promising syntheic pathway for chemists;
2) chemists verify the route quality and determine the necessary actions for robotic arms;
3) the robotic arm completes the entire synthesis process according to the configured operating procedures. The advent of ASKCOS represents a great advancement in CASP, significantly reducing the workload of chemists on manual tasks. Now the authors have deployed the first step of retrosynthesis automatic planning on the website~\footnote{https://askcos.mit.edu/} and allowed free access.

Genheden~\textit{et al.}~\cite{genheden2020aizynthfinder} developed a another open-source CASP software AiZynthFinder, which works similarly to ASKCOS but more efficiently.  They aim to provide a robust and transparent platform that not only reduces the learning cost for new users, but also facilitates researchers and integrates more retrosynthesis algorithms in the future. Now it has implemented various multi-step search algorithms, including DFPN, MCTS, and Retro*. The entire software is deployed on a python GUI rather than a web page.

IBM RoboRXN is a reaction prediction tool built on the works of Schwaller~\textit{et al.}~\cite{schwaller2019molecular,schwaller2020predicting}. Deployed on the website~\footnote{https://rxn.res.ibm.com/}, it provides the friendly interface for users to perform both forward reaction and retrosynthesis route prediction.  After receiving the molecular structure drawn by the user or SMILES, the website will provide the corresponding prediction results along with the confidence. The attention weight map between the product SMILES and reactant SMILES is also available. Moreover, for retrosynthesis predictions, it allows the user to add some additional constraints and provides multiple optional results at the same time. 

\subsubsection{Commercial Platform}
Most of commercial platforms are closed where algorithms, hand-coded reaction rules, or databases are unavailable.
Here we can only give some short descriptions for them.
Synthia~\cite{mikulak2020computational}, aka Chematica, is a hybrid expert-AI system that contains over 100, 000 reaction rules manually coded by chemists and also adopted machine learning algorithms. 
As mature commercial platforms, Reaxys synthesis planning~\footnote{https://www.elsevier.com/en-xm/solutions/reaxys/how-reaxys-works/synthesis-planner}, CAS SciFinder$^n$~\footnote{https://www.cas.org/solutions/cas-scifinder-discovery-platform/cas-scifinder}, and ChemAIRS~\footnote{https://www.chemical.ai/} all can simultaneously provide reaction conditions, synthetic routes, and real-time purchase advice for target compounds.

Similar to the double-blind AB tests conducted by Selger~\textit{et al.}~\cite{segler2018planning}, these commercial platforms were also compared with experienced chemists.
The latest research of Synthia~\cite{mikulak2020computational} demonstrates that it has passed the Turing test, where a chemist cannot tell whether a synthetic route was designed by an experienced chemist or an AI system. Apart from that, the developes of ChemAIRS conducted a meaningful competition between ChemAIRS and 16 experienced chemists. They both were asked to propose synthetic routes for 22 molecules without chirality whose synthetic steps range from 8 to 14. The results of the competition shows that not only is the route quality of the ChemAIRS comparable to that of the chemist, but also it outperforms the chemists in terms of design speed and route diversity. 
All of the above studies demonstrates that the current CASP method has reached a level comparable to that of experienced chemists.

\section{Challenges and Opportunities}
\subsection{More Reasonable Metric}
In our opinion, the current single-step retrosynthesis tasks, both single-step and multi-step, lack more reasonable and universal metrics in realistic scenarios.
While the top-$K$ accuracy and success rate successfully measure whether the model has mastered the ability to predict precursors or synthetic routes, they are incapable of judging whether the model could predict new feasible ones. 
Since it is unknown whether a synthetic proposal is chemically feasible until the experiment is performed in the laboratory, the model should also focus on low cost~\cite{badowski2019selection}, high-yielding, and diversity~\cite{chen2019learning,wang2021retroprime,genheden2022paroutes}.
However, indicators for these aspects are relatively few and are not given enough attention.
Inspired by the reference model used by Kim~\textit{et al.}~\cite{kim2021self} that can eliminate unrealistic reactions, it may be possible to train a new neural network model to determine whether a reaction or reaction route meets the above requirements with existing reaction databases.


There is another problem for multi-step retrosynthesis. The current value function is built with the aim of finding the shortest path in mind. Thus, it naturally loses sight of the factors that make the path longer but necessary~(for example, the protection and deprotection strategies are incompatible). In short, it is difficult for the algorithm to reproduce the complete synthetic route in the literature due to its design principle.

\subsection{More Data}
The currently available data is far from adequate in terms of both volume and diversity.
First of all, the amount of data in the current public datasets is far from sufficient. While the number of reactions in current public datasets is around 1 million, the number of molecules with property annotations in the ZINC~\cite{irwin2020zinc20} database is over 1.4 billion.
In addition, current datasets usually contain only the successful reactions rather than the failed ones. These failed reaction data are also crucial for model training~\cite{kurczab2014influence}.
Finally, reaction conditions such as temperature and catalyst are always ignored for the existing methods, which could be critical to the feasibility of the reaction in some cases. Theoretically, these reaction conditions can guide the model predictions to some extent and serve as the predicted objects to help chemists discover feasible reaction conditions. The collection of reaction databases still has a long way to go.


\subsection{Interpretability}
A significant problem for existing AI-based retrosynthesis is the lack of interpretability.
A recent study has shown that machine learning models for chemistry applications may sometimes simply capture literature popularity trends, as illustrated by the example of predicting reactions for heterocyclic Suzuki–Miyaura coupling~\cite{beker2022machine}. 
An outstanding organic synthesis technique should not only solve existing synthesis problems but also inspire chemists to design better drug molecules~\cite{campos2019importance}, which makes it essential to allow chemists to gain insights from models.
Current retrosynthesis work tends to articulate interpretability in the goal of retrosynthesis itself. For instance, some template-based methods consider the reaction template to be interpretable, and semi-template methods consider synthons to be interpretable.
While this approach interprets the retrosynthesis to some extent, it does not explain how the model makes this prediction.  

Some studies mentioned in previous sections have already made notable progress in this direction.
In particular, some Transformer-based methods~\cite{sumner2020levenshtein,zhong2022root} use the cross-attention mechanism in Transformer to illustrate the relationship between reactant tokens and resultant tokens, which well explains why most substructures in the product are kept in the reactants. However, it does not account for any new atoms and bonds in the reactants. 
The interpretable retrosynthesis method should be able to attribute predictions to specific atoms or substructures.
Some work has been conducted on attributing properties to molecular fragments for molecular property prediction tasks~\cite{xu2017deep,wu2021mining,jia2022explainable}. 
If a similar attribution could be achieved in retrosynthesis, it would greatly help researchers to design drug molecules~\cite{murray2009rise}.

 
\section{Conclusion}
Over the past few years, motivated by both the development of AI and industrial demands, AI-based retrosynthesis has become an inspiring research area. Due to its automatic extraction and learning of reaction principles, it significantly outperforms traditional expert systems based on manual extraction in terms of cost and efficiency. For both single and multi-step, we can see dramatic progress in academia and some mature commercial applications, which fully demonstrate the great promise of this filed.

However, despite the tremendous progress in recent years, the area of retrosynthesis driven by artificial intelligence is far from a mature state. 
The ultimate goal of retrosynthesis is to discover synthetic routes for new drugs or to discover new synthetic routes for existing drugs with a lower cost and higher yield.
However, currently we have not seen a study reporting a brand-new synthetic route that was provided entirely by artificial intelligence and successfully synthesized in the laboratory or industrial setting.
To report a plausible and novel synthetic route, it may be necessary for chemists and computer experts to cooperate with each other closely.
As more and more researchers are placed in the field of retrosynthesis, we believe that related studies will come out soon.

\bibliography{sn-bibliography}

\begin{thebibliography}{100}
\expandafter\ifx\csname url\endcsname\relax
  \def\url#1{\burl{#1}}\fi
\expandafter\ifx\csname urlprefix\endcsname\relax\def\urlprefix{}\fi
\providecommand{\bibinfo}[2]{#2}
\providecommand{\eprint}[2][]{\url{#2}}
\providecommand{\doi}[1]{\url{https://doi.org/#1}}
\bibcommenthead

\bibitem{wouters2020estimated}
\bibinfo{author}{Wouters, O.~J.}, \bibinfo{author}{McKee, M.} \&
  \bibinfo{author}{Luyten, J.}
\newblock \bibinfo{title}{Estimated research and development investment needed
  to bring a new medicine to market, 2009-2018}.
\newblock \emph{\bibinfo{journal}{Jama}} \textbf{\bibinfo{volume}{323}}~(9),
  \bibinfo{pages}{844--853} (\bibinfo{year}{2020}).

\bibitem{blakemore2018organic}
\bibinfo{author}{Blakemore, D.~C.} \emph{et~al.}
\newblock \bibinfo{title}{Organic synthesis provides opportunities to transform
  drug discovery}.
\newblock \emph{\bibinfo{journal}{Nature chemistry}}
  \textbf{\bibinfo{volume}{10}}~(4), \bibinfo{pages}{383--394}
  (\bibinfo{year}{2018}).

\bibitem{campos2019importance}
\bibinfo{author}{Campos, K.~R.} \emph{et~al.}
\newblock \bibinfo{title}{The importance of synthetic chemistry in the
  pharmaceutical industry}.
\newblock \emph{\bibinfo{journal}{Science}}
  \textbf{\bibinfo{volume}{363}}~(6424), \bibinfo{pages}{eaat0805}
  (\bibinfo{year}{2019}).

\bibitem{corey1991multistep}
\bibinfo{author}{Corey, E.~J.}
\newblock \bibinfo{title}{The logic of chemical synthesis: multistep synthesis
  of complex carbogenic molecules (nobel lecture)}.
\newblock \emph{\bibinfo{journal}{Angewandte Chemie International Edition in
  English}} \textbf{\bibinfo{volume}{30}}~(5), \bibinfo{pages}{455--465}
  (\bibinfo{year}{1991}).

\bibitem{corey1988robert}
\bibinfo{author}{Corey, E.~J.}
\newblock \bibinfo{title}{Robert robinson lecture. retrosynthetic
  thinking—essentials and examples}.
\newblock \emph{\bibinfo{journal}{Chemical society reviews}}
  \textbf{\bibinfo{volume}{17}}, \bibinfo{pages}{111--133}
  (\bibinfo{year}{1988}).

\bibitem{rogers2010extended}
\bibinfo{author}{Rogers, D.} \& \bibinfo{author}{Hahn, M.}
\newblock \bibinfo{title}{Extended-connectivity fingerprints}.
\newblock \emph{\bibinfo{journal}{Journal of chemical information and
  modeling}} \textbf{\bibinfo{volume}{50}}~(5), \bibinfo{pages}{742--754}
  (\bibinfo{year}{2010}).

\bibitem{todeschini2008handbook}
\bibinfo{author}{Todeschini, R.} \& \bibinfo{author}{Consonni, V.}
\newblock \emph{\bibinfo{title}{Handbook of molecular descriptors}}
  (\bibinfo{publisher}{John Wiley \& Sons}, \bibinfo{year}{2008}).

\bibitem{david2020molecular}
\bibinfo{author}{David, L.}, \bibinfo{author}{Thakkar, A.},
  \bibinfo{author}{Mercado, R.} \& \bibinfo{author}{Engkvist, O.}
\newblock \bibinfo{title}{Molecular representations in ai-driven drug
  discovery: a review and practical guide}.
\newblock \emph{\bibinfo{journal}{Journal of Cheminformatics}}
  \textbf{\bibinfo{volume}{12}}~(1), \bibinfo{pages}{1--22}
  (\bibinfo{year}{2020}).

\bibitem{vaswani2017attention}
\bibinfo{author}{Vaswani, A.} \emph{et~al.}
\newblock \bibinfo{title}{Attention is all you need}.
\newblock \emph{\bibinfo{journal}{Advances in neural information processing
  systems}} \textbf{\bibinfo{volume}{30}} (\bibinfo{year}{2017}).

\bibitem{szymkuc2016computer}
\bibinfo{author}{Szymku{\'c}, S.} \emph{et~al.}
\newblock \bibinfo{title}{Computer-assisted synthetic planning: the end of the
  beginning}.
\newblock \emph{\bibinfo{journal}{Angewandte Chemie International Edition}}
  \textbf{\bibinfo{volume}{55}}~(20), \bibinfo{pages}{5904--5937}
  (\bibinfo{year}{2016}).

\bibitem{pensak1977lhasa}
\bibinfo{author}{Pensak, D.~A.} \& \bibinfo{author}{Corey, E.~J.}
\newblock \emph{\bibinfo{title}{LHASA—logic and heuristics applied to
  synthetic analysis}}  (\bibinfo{publisher}{ACS Publications},
  \bibinfo{year}{1977}).

\bibitem{johnson1989designing}
\bibinfo{author}{Johnson, P.} \emph{et~al.}
\newblock \bibinfo{title}{Designing an expert system for organic synthesis in
  expert systems application in chemistry}.
\newblock In \emph{\bibinfo{booktitle}{ACS Symposium Series of American
  Chemical Society}}  (\bibinfo{year}{1989}).

\bibitem{gasteiger2000computer}
\bibinfo{author}{Gasteiger, J.} \emph{et~al.}
\newblock \bibinfo{title}{Computer-assisted synthesis and reaction planning in
  combinatorial chemistry}.
\newblock \emph{\bibinfo{journal}{Perspectives in Drug Discovery and Design}}
  \textbf{\bibinfo{volume}{20}}~(1), \bibinfo{pages}{245--264}
  (\bibinfo{year}{2000}).

\bibitem{feng2018computational}
\bibinfo{author}{Feng, F.}, \bibinfo{author}{Lai, L.} \& \bibinfo{author}{Pei,
  J.}
\newblock \bibinfo{title}{Computational chemical synthesis analysis and pathway
  design}.
\newblock \emph{\bibinfo{journal}{Frontiers in chemistry}}
  \textbf{\bibinfo{volume}{6}}, \bibinfo{pages}{199} (\bibinfo{year}{2018}).

\bibitem{chen2016evolution}
\bibinfo{author}{Chen, J.~X.}
\newblock \bibinfo{title}{The evolution of computing: Alphago}.
\newblock \emph{\bibinfo{journal}{Computing in Science \& Engineering}}
  \textbf{\bibinfo{volume}{18}}~(4), \bibinfo{pages}{4--7}
  (\bibinfo{year}{2016}).

\bibitem{cai2020review}
\bibinfo{author}{Cai, L.}, \bibinfo{author}{Gao, J.} \& \bibinfo{author}{Zhao,
  D.}
\newblock \bibinfo{title}{A review of the application of deep learning in
  medical image classification and segmentation}.
\newblock \emph{\bibinfo{journal}{Annals of translational medicine}}
  \textbf{\bibinfo{volume}{8}}~(11) (\bibinfo{year}{2020}).

\bibitem{guo2020systematic}
\bibinfo{author}{Guo, X.} \& \bibinfo{author}{Zhao, L.}
\newblock \bibinfo{title}{A systematic survey on deep generative models for
  graph generation} (\bibinfo{year}{2020}).
\newblock \bibinfo{note}{Preprint at \url{https://arxiv.org/abs/2007.06686}}.

\bibitem{baskin2016renaissance}
\bibinfo{author}{Baskin, I.~I.}, \bibinfo{author}{Winkler, D.} \&
  \bibinfo{author}{Tetko, I.~V.}
\newblock \bibinfo{title}{A renaissance of neural networks in drug discovery}.
\newblock \emph{\bibinfo{journal}{Expert opinion on drug discovery}}
  \textbf{\bibinfo{volume}{11}}~(8), \bibinfo{pages}{785--795}
  (\bibinfo{year}{2016}).

\bibitem{chen2018rise}
\bibinfo{author}{Chen, H.}, \bibinfo{author}{Engkvist, O.},
  \bibinfo{author}{Wang, Y.}, \bibinfo{author}{Olivecrona, M.} \&
  \bibinfo{author}{Blaschke, T.}
\newblock \bibinfo{title}{The rise of deep learning in drug discovery}.
\newblock \emph{\bibinfo{journal}{Drug discovery today}}
  \textbf{\bibinfo{volume}{23}}~(6), \bibinfo{pages}{1241--1250}
  (\bibinfo{year}{2018}).

\bibitem{heifets2012construction}
\bibinfo{author}{Heifets, A.} \& \bibinfo{author}{Jurisica, I.}
\newblock \bibinfo{title}{Construction of new medicines via game proof search}.
\newblock In \emph{\bibinfo{booktitle}{Proceedings of the AAAI Conference on
  Artificial Intelligence}} \bibinfo{pages}{1564--1570} (\bibinfo{year}{2012}).

\bibitem{segler2017neural}
\bibinfo{author}{Segler, M.~H.} \& \bibinfo{author}{Waller, M.~P.}
\newblock \bibinfo{title}{Neural-symbolic machine learning for retrosynthesis
  and reaction prediction}.
\newblock \emph{\bibinfo{journal}{Chemistry--A European Journal}}
  \textbf{\bibinfo{volume}{23}}~(25), \bibinfo{pages}{5966--5971}
  (\bibinfo{year}{2017}).

\bibitem{yi2016study}
\bibinfo{author}{Yi, H.}, \bibinfo{author}{Shiyu, S.},
  \bibinfo{author}{Xiusheng, D.} \& \bibinfo{author}{Zhigang, C.}
\newblock \bibinfo{title}{A study on deep neural networks framework}.
\newblock In \emph{\bibinfo{booktitle}{2016 IEEE Advanced Information
  Management, Communicates, Electronic and Automation Control Conference
  (IMCEC)}} \bibinfo{pages}{1519--1522} (\bibinfo{organization}{IEEE},
  \bibinfo{year}{2016}).

\bibitem{browne2012mcts}
\bibinfo{author}{Browne, C.~B.} \emph{et~al.}
\newblock \bibinfo{title}{A survey of monte carlo tree search methods}.
\newblock \emph{\bibinfo{journal}{IEEE Transactions on Computational
  Intelligence and AI in games}} \textbf{\bibinfo{volume}{4}}~(1),
  \bibinfo{pages}{1--43} (\bibinfo{year}{2012}).

\bibitem{segler2018planning}
\bibinfo{author}{Segler, M.~H.}, \bibinfo{author}{Preuss, M.} \&
  \bibinfo{author}{Waller, M.~P.}
\newblock \bibinfo{title}{Planning chemical syntheses with deep neural networks
  and symbolic ai}.
\newblock \emph{\bibinfo{journal}{Nature}}
  \textbf{\bibinfo{volume}{555}}~(7698), \bibinfo{pages}{604--610}
  (\bibinfo{year}{2018}).

\bibitem{guo2020bayesian}
\bibinfo{author}{Guo, Z.}, \bibinfo{author}{Wu, S.}, \bibinfo{author}{Ohno, M.}
  \& \bibinfo{author}{Yoshida, R.}
\newblock \bibinfo{title}{Bayesian algorithm for retrosynthesis}.
\newblock \emph{\bibinfo{journal}{Journal of Chemical Information and
  Modeling}} \textbf{\bibinfo{volume}{60}}~(10), \bibinfo{pages}{4474--4486}
  (\bibinfo{year}{2020}).

\bibitem{lee2021retcl}
\bibinfo{author}{Lee, H.} \emph{et~al.}
\newblock \bibinfo{title}{Retcl: A selection-based approach for retrosynthesis
  via contrastive learning}.
\newblock In \emph{\bibinfo{booktitle}{Proceedings of the Thirtieth
  International Joint Conference on Artificial Intelligence}}
  (\bibinfo{year}{2021}).

\bibitem{fortunato2020data}
\bibinfo{author}{Fortunato, M.~E.}, \bibinfo{author}{Coley, C.~W.},
  \bibinfo{author}{Barnes, B.~C.} \& \bibinfo{author}{Jensen, K.~F.}
\newblock \bibinfo{title}{Data augmentation and pretraining for template-based
  retrosynthetic prediction in computer-aided synthesis planning}.
\newblock \emph{\bibinfo{journal}{Journal of chemical information and
  modeling}} \textbf{\bibinfo{volume}{60}}~(7), \bibinfo{pages}{3398--3407}
  (\bibinfo{year}{2020}).

\bibitem{coley2017computer}
\bibinfo{author}{Coley, C.~W.}, \bibinfo{author}{Rogers, L.},
  \bibinfo{author}{Green, W.~H.} \& \bibinfo{author}{Jensen, K.~F.}
\newblock \bibinfo{title}{Computer-assisted retrosynthesis based on molecular
  similarity}.
\newblock \emph{\bibinfo{journal}{ACS central science}}
  \textbf{\bibinfo{volume}{3}}~(12), \bibinfo{pages}{1237--1245}
  (\bibinfo{year}{2017}).

\bibitem{seidl2022improving}
\bibinfo{author}{Seidl, P.} \emph{et~al.}
\newblock \bibinfo{title}{Improving few-and zero-shot reaction template
  prediction using modern hopfield networks}.
\newblock \emph{\bibinfo{journal}{Journal of chemical information and
  modeling}}  (\bibinfo{year}{2022}).

\bibitem{ishida2019prediction}
\bibinfo{author}{Ishida, S.}, \bibinfo{author}{Terayama, K.},
  \bibinfo{author}{Kojima, R.}, \bibinfo{author}{Takasu, K.} \&
  \bibinfo{author}{Okuno, Y.}
\newblock \bibinfo{title}{Prediction and interpretable visualization of
  retrosynthetic reactions using graph convolutional networks}.
\newblock \emph{\bibinfo{journal}{Journal of chemical information and
  modeling}} \textbf{\bibinfo{volume}{59}}~(12), \bibinfo{pages}{5026--5033}
  (\bibinfo{year}{2019}).

\bibitem{dai2019retrosynthesis}
\bibinfo{author}{Dai, H.}, \bibinfo{author}{Li, C.}, \bibinfo{author}{Coley,
  C.}, \bibinfo{author}{Dai, B.} \& \bibinfo{author}{Song, L.}
\newblock \bibinfo{title}{Retrosynthesis prediction with conditional graph
  logic network}.
\newblock \emph{\bibinfo{journal}{Advances in Neural Information Processing
  Systems}} \textbf{\bibinfo{volume}{32}}, \bibinfo{pages}{8872--8882}
  (\bibinfo{year}{2019}).

\bibitem{chen2021deep}
\bibinfo{author}{Chen, S.} \& \bibinfo{author}{Jung, Y.}
\newblock \bibinfo{title}{Deep retrosynthetic reaction prediction using local
  reactivity and global attention}.
\newblock \emph{\bibinfo{journal}{JACS Au}} \textbf{\bibinfo{volume}{1}}~(10),
  \bibinfo{pages}{1612--1620} (\bibinfo{year}{2021}).

\bibitem{wang2021retroprime}
\bibinfo{author}{Wang, X.} \emph{et~al.}
\newblock \bibinfo{title}{Retroprime: A diverse, plausible and
  transformer-based method for single-step retrosynthesis predictions}.
\newblock \emph{\bibinfo{journal}{Chemical Engineering Journal}}
  \textbf{\bibinfo{volume}{420}}, \bibinfo{pages}{129845}
  (\bibinfo{year}{2021}).

\bibitem{shi2020graph}
\bibinfo{author}{Shi, C.}, \bibinfo{author}{Xu, M.}, \bibinfo{author}{Guo, H.},
  \bibinfo{author}{Zhang, M.} \& \bibinfo{author}{Tang, J.}
\newblock \bibinfo{title}{A graph to graphs framework for retrosynthesis
  prediction}.
\newblock In \emph{\bibinfo{booktitle}{International Conference on Machine
  Learning}} \bibinfo{pages}{8818--8827} (\bibinfo{organization}{PMLR},
  \bibinfo{year}{2020}).

\bibitem{somnath2021learning}
\bibinfo{author}{Somnath, V.~R.}, \bibinfo{author}{Bunne, C.},
  \bibinfo{author}{Coley, C.~W.}, \bibinfo{author}{Krause, A.} \&
  \bibinfo{author}{Barzilay, R.}
\newblock \bibinfo{title}{Learning graph models for retrosynthesis prediction}.
\newblock In \emph{\bibinfo{booktitle}{Thirty-Fifth Conference on Neural
  Information Processing Systems}}  (\bibinfo{year}{2021}).

\bibitem{yan2020retroxpert}
\bibinfo{author}{Yan, C.} \emph{et~al.}
\newblock \bibinfo{title}{Retroxpert: Decompose retrosynthesis prediction like
  a chemist}.
\newblock \emph{\bibinfo{journal}{Advances in Neural Information Processing
  Systems}} \textbf{\bibinfo{volume}{33}}, \bibinfo{pages}{11248--11258}
  (\bibinfo{year}{2020}).

\bibitem{liu2017retrosynthetic}
\bibinfo{author}{Liu, B.} \emph{et~al.}
\newblock \bibinfo{title}{Retrosynthetic reaction prediction using neural
  sequence-to-sequence models}.
\newblock \emph{\bibinfo{journal}{ACS central science}}
  \textbf{\bibinfo{volume}{3}}~(10), \bibinfo{pages}{1103--1113}
  (\bibinfo{year}{2017}).

\bibitem{karpov2019transformer}
\bibinfo{author}{Karpov, P.}, \bibinfo{author}{Godin, G.} \&
  \bibinfo{author}{Tetko, I.~V.}
\newblock \bibinfo{title}{A transformer model for retrosynthesis}.
\newblock In \emph{\bibinfo{booktitle}{International Conference on Artificial
  Neural Networks}} \bibinfo{pages}{817--830}
  (\bibinfo{organization}{Springer}, \bibinfo{year}{2019}).

\bibitem{zheng2019predicting}
\bibinfo{author}{Zheng, S.}, \bibinfo{author}{Rao, J.}, \bibinfo{author}{Zhang,
  Z.}, \bibinfo{author}{Xu, J.} \& \bibinfo{author}{Yang, Y.}
\newblock \bibinfo{title}{Predicting retrosynthetic reactions using
  self-corrected transformer neural networks}.
\newblock \emph{\bibinfo{journal}{Journal of Chemical Information and
  Modeling}} \textbf{\bibinfo{volume}{60}}~(1), \bibinfo{pages}{47--55}
  (\bibinfo{year}{2019}).

\bibitem{chen2019learning}
\bibinfo{author}{Chen, B.}, \bibinfo{author}{Shen, T.},
  \bibinfo{author}{Jaakkola, T.~S.} \& \bibinfo{author}{Barzilay, R.}
\newblock \bibinfo{title}{Learning to make generalizable and diverse
  predictions for retrosynthesis} (\bibinfo{year}{2019}).
\newblock \bibinfo{note}{Preprint at \url{https://arxiv.org/abs/1910.09688}}.

\bibitem{yang2019molecular}
\bibinfo{author}{Yang, Q.} \emph{et~al.}
\newblock \bibinfo{title}{Molecular transformer unifies reaction prediction and
  retrosynthesis across pharma chemical space}.
\newblock \emph{\bibinfo{journal}{Chemical communications}}
  \textbf{\bibinfo{volume}{55}}~(81), \bibinfo{pages}{12152--12155}
  (\bibinfo{year}{2019}).

\bibitem{lin2020automatic}
\bibinfo{author}{Lin, K.}, \bibinfo{author}{Xu, Y.}, \bibinfo{author}{Pei, J.}
  \& \bibinfo{author}{Lai, L.}
\newblock \bibinfo{title}{Automatic retrosynthetic route planning using
  template-free models}.
\newblock \emph{\bibinfo{journal}{Chemical science}}
  \textbf{\bibinfo{volume}{11}}~(12), \bibinfo{pages}{3355--3364}
  (\bibinfo{year}{2020}).

\bibitem{tetko2020state}
\bibinfo{author}{Tetko, I.~V.}, \bibinfo{author}{Karpov, P.},
  \bibinfo{author}{Van~Deursen, R.} \& \bibinfo{author}{Godin, G.}
\newblock \bibinfo{title}{State-of-the-art augmented nlp transformer models for
  direct and single-step retrosynthesis}.
\newblock \emph{\bibinfo{journal}{Nature communications}}
  \textbf{\bibinfo{volume}{11}}~(1), \bibinfo{pages}{1--11}
  (\bibinfo{year}{2020}).

\bibitem{seo2021gta}
\bibinfo{author}{Seo, S.-W.} \emph{et~al.}
\newblock \bibinfo{title}{Gta: Graph truncated attention for retrosynthesis}.
\newblock \emph{\bibinfo{journal}{Proceedings of the AAAI Conference on
  Artificial Intelligence}} \textbf{\bibinfo{volume}{35}}~(1),
  \bibinfo{pages}{531--539} (\bibinfo{year}{2021}).

\bibitem{kim2021valid}
\bibinfo{author}{Kim, E.}, \bibinfo{author}{Lee, D.}, \bibinfo{author}{Kwon,
  Y.}, \bibinfo{author}{Park, M.~S.} \& \bibinfo{author}{Choi, Y.-S.}
\newblock \bibinfo{title}{Valid, plausible, and diverse retrosynthesis using
  tied two-way transformers with latent variables}.
\newblock \emph{\bibinfo{journal}{Journal of Chemical Information and
  Modeling}} \textbf{\bibinfo{volume}{61}}~(1), \bibinfo{pages}{123--133}
  (\bibinfo{year}{2021}).

\bibitem{irwin2022chemformer}
\bibinfo{author}{Irwin, R.}, \bibinfo{author}{Dimitriadis, S.},
  \bibinfo{author}{He, J.} \& \bibinfo{author}{Bjerrum, E.~J.}
\newblock \bibinfo{title}{Chemformer: a pre-trained transformer for
  computational chemistry}.
\newblock \emph{\bibinfo{journal}{Machine Learning: Science and Technology}}
  \textbf{\bibinfo{volume}{3}}~(1), \bibinfo{pages}{015022}
  (\bibinfo{year}{2022}).

\bibitem{zhong2022root}
\bibinfo{author}{Zhong, Z.} \emph{et~al.}
\newblock \bibinfo{title}{Root-aligned smiles: a tight representation for
  chemical reaction prediction}.
\newblock \emph{\bibinfo{journal}{Chemical Science}}
  \textbf{\bibinfo{volume}{13}}~(31), \bibinfo{pages}{9023--9034}
  (\bibinfo{year}{2022}).

\bibitem{sacha2021molecule}
\bibinfo{author}{Sacha, M.} \emph{et~al.}
\newblock \bibinfo{title}{Molecule edit graph attention network: modeling
  chemical reactions as sequences of graph edits}.
\newblock \emph{\bibinfo{journal}{Journal of Chemical Information and
  Modeling}} \textbf{\bibinfo{volume}{61}}~(7), \bibinfo{pages}{3273--3284}
  (\bibinfo{year}{2021}).

\bibitem{mao2021molecular}
\bibinfo{author}{Mao, K.} \emph{et~al.}
\newblock \bibinfo{title}{Molecular graph enhanced transformer for
  retrosynthesis prediction}.
\newblock \emph{\bibinfo{journal}{Neurocomputing}}
  \textbf{\bibinfo{volume}{457}}, \bibinfo{pages}{193--202}
  (\bibinfo{year}{2021}).

\bibitem{mann2021retrosynthesis}
\bibinfo{author}{Mann, V.} \& \bibinfo{author}{Venkatasubramanian, V.}
\newblock \bibinfo{title}{Retrosynthesis prediction using grammar-based neural
  machine translation: An information-theoretic approach}.
\newblock \emph{\bibinfo{journal}{Computers \& Chemical Engineering}}
  \textbf{\bibinfo{volume}{155}}, \bibinfo{pages}{107533}
  (\bibinfo{year}{2021}).

\bibitem{ucak2021substructure}
\bibinfo{author}{Ucak, U.~V.}, \bibinfo{author}{Kang, T.}, \bibinfo{author}{Ko,
  J.} \& \bibinfo{author}{Lee, J.}
\newblock \bibinfo{title}{Substructure-based neural machine translation for
  retrosynthetic prediction}.
\newblock \emph{\bibinfo{journal}{Journal of cheminformatics}}
  \textbf{\bibinfo{volume}{13}}~(1), \bibinfo{pages}{1--15}
  (\bibinfo{year}{2021}).

\bibitem{ucak2022retrosynthetic}
\bibinfo{author}{Ucak, U.~V.}, \bibinfo{author}{Ashyrmamatov, I.},
  \bibinfo{author}{Ko, J.} \& \bibinfo{author}{Lee, J.}
\newblock \bibinfo{title}{Retrosynthetic reaction pathway prediction through
  neural machine translation of atomic environments}.
\newblock \emph{\bibinfo{journal}{Nature communications}}
  \textbf{\bibinfo{volume}{13}}~(1), \bibinfo{pages}{1--10}
  (\bibinfo{year}{2022}).

\bibitem{schwaller2019molecular}
\bibinfo{author}{Schwaller, P.} \emph{et~al.}
\newblock \bibinfo{title}{Molecular transformer: a model for
  uncertainty-calibrated chemical reaction prediction}.
\newblock \emph{\bibinfo{journal}{ACS central science}}
  \textbf{\bibinfo{volume}{5}}~(9), \bibinfo{pages}{1572--1583}
  (\bibinfo{year}{2019}).

\bibitem{coley2019rdchiral}
\bibinfo{author}{Coley, C.~W.}, \bibinfo{author}{Green, W.~H.} \&
  \bibinfo{author}{Jensen, K.~F.}
\newblock \bibinfo{title}{Rdchiral: An rdkit wrapper for handling
  stereochemistry in retrosynthetic template extraction and application}.
\newblock \emph{\bibinfo{journal}{Journal of chemical information and
  modeling}} \textbf{\bibinfo{volume}{59}}~(6), \bibinfo{pages}{2529--2537}
  (\bibinfo{year}{2019}).

\bibitem{srivastava2015highway}
\bibinfo{author}{Srivastava, R.~K.}, \bibinfo{author}{Greff, K.} \&
  \bibinfo{author}{Schmidhuber, J.}
\newblock \bibinfo{title}{Highway networks} (\bibinfo{year}{2015}).
\newblock \bibinfo{note}{Preprint at \url{https://arxiv.org/abs/1505.00387}}.

\bibitem{ramsauer2020hopfield}
\bibinfo{author}{Ramsauer, H.} \emph{et~al.}
\newblock \bibinfo{title}{Hopfield networks is all you need}
  (\bibinfo{year}{2020}).
\newblock \bibinfo{note}{Preprint at \url{https://arxiv.org/abs/2008.02217}}.

\bibitem{widrich2020modern}
\bibinfo{author}{Widrich, M.} \emph{et~al.}
\newblock \bibinfo{title}{Modern hopfield networks and attention for immune
  repertoire classification}.
\newblock \emph{\bibinfo{journal}{Advances in Neural Information Processing
  Systems}} \textbf{\bibinfo{volume}{33}}, \bibinfo{pages}{18832--18845}
  (\bibinfo{year}{2020}).

\bibitem{kipf2016semi}
\bibinfo{author}{Kipf, T.~N.} \& \bibinfo{author}{Welling, M.}
\newblock \bibinfo{title}{Semi-supervised classification with graph
  convolutional networks} (\bibinfo{year}{2016}).
\newblock \bibinfo{note}{Preprint at \url{https://arxiv.org/abs/1609.02907}}.

\bibitem{sundararajan2017axiomatic}
\bibinfo{author}{Sundararajan, M.}, \bibinfo{author}{Taly, A.} \&
  \bibinfo{author}{Yan, Q.}
\newblock \bibinfo{title}{Axiomatic attribution for deep networks}.
\newblock In \emph{\bibinfo{booktitle}{International conference on machine
  learning}} \bibinfo{pages}{3319--3328} (\bibinfo{organization}{PMLR},
  \bibinfo{year}{2017}).

\bibitem{gilmer2017neural}
\bibinfo{author}{Gilmer, J.}, \bibinfo{author}{Schoenholz, S.~S.},
  \bibinfo{author}{Riley, P.~F.}, \bibinfo{author}{Vinyals, O.} \&
  \bibinfo{author}{Dahl, G.~E.}
\newblock \bibinfo{title}{Neural message passing for quantum chemistry}.
\newblock In \emph{\bibinfo{booktitle}{International conference on machine
  learning}} \bibinfo{pages}{1263--1272} (\bibinfo{organization}{PMLR},
  \bibinfo{year}{2017}).

\bibitem{daud2020applications}
\bibinfo{author}{Daud, N.~N.}, \bibinfo{author}{Ab~Hamid, S.~H.},
  \bibinfo{author}{Saadoon, M.}, \bibinfo{author}{Sahran, F.} \&
  \bibinfo{author}{Anuar, N.~B.}
\newblock \bibinfo{title}{Applications of link prediction in social networks: A
  review}.
\newblock \emph{\bibinfo{journal}{Journal of Network and Computer
  Applications}} \textbf{\bibinfo{volume}{166}}, \bibinfo{pages}{102716}
  (\bibinfo{year}{2020}).

\bibitem{rossi2021knowledge}
\bibinfo{author}{Rossi, A.}, \bibinfo{author}{Barbosa, D.},
  \bibinfo{author}{Firmani, D.}, \bibinfo{author}{Matinata, A.} \&
  \bibinfo{author}{Merialdo, P.}
\newblock \bibinfo{title}{Knowledge graph embedding for link prediction: A
  comparative analysis}.
\newblock \emph{\bibinfo{journal}{ACM Transactions on Knowledge Discovery from
  Data (TKDD)}} \textbf{\bibinfo{volume}{15}}~(2), \bibinfo{pages}{1--49}
  (\bibinfo{year}{2021}).

\bibitem{schlichtkrull2018modeling}
\bibinfo{author}{Schlichtkrull, M.} \emph{et~al.}
\newblock \bibinfo{title}{Modeling relational data with graph convolutional
  networks}.
\newblock In \emph{\bibinfo{booktitle}{European semantic web conference}}
  \bibinfo{pages}{593--607} (\bibinfo{organization}{Springer},
  \bibinfo{year}{2018}).

\bibitem{puterman1990markov}
\bibinfo{author}{Puterman, M.~L.}
\newblock \bibinfo{title}{Markov decision processes}.
\newblock \emph{\bibinfo{journal}{Handbooks in operations research and
  management science}} \textbf{\bibinfo{volume}{2}}, \bibinfo{pages}{331--434}
  (\bibinfo{year}{1990}).

\bibitem{velickovic2017graph}
\bibinfo{author}{Velickovic, P.} \emph{et~al.}
\newblock \bibinfo{title}{Graph attention networks} (\bibinfo{year}{2017}).
\newblock \bibinfo{note}{Preprint at \url{https://arxiv.org/abs/1710.10903}}.

\bibitem{yang2020survey}
\bibinfo{author}{Yang, S.}, \bibinfo{author}{Wang, Y.} \& \bibinfo{author}{Chu,
  X.}
\newblock \bibinfo{title}{A survey of deep learning techniques for neural
  machine translation} (\bibinfo{year}{2020}).
\newblock \bibinfo{note}{Preprint at \url{https://arxiv.org/abs/2002.07526}}.

\bibitem{van2020review}
\bibinfo{author}{Van~Houdt, G.}, \bibinfo{author}{Mosquera, C.} \&
  \bibinfo{author}{N{\'a}poles, G.}
\newblock \bibinfo{title}{A review on the long short-term memory model}.
\newblock \emph{\bibinfo{journal}{Artificial Intelligence Review}}
  \textbf{\bibinfo{volume}{53}}~(8), \bibinfo{pages}{5929--5955}
  (\bibinfo{year}{2020}).

\bibitem{parikh2016decomposable}
\bibinfo{author}{Parikh, A.~P.}, \bibinfo{author}{T{\"a}ckstr{\"o}m, O.},
  \bibinfo{author}{Das, D.} \& \bibinfo{author}{Uszkoreit, J.}
\newblock \bibinfo{title}{A decomposable attention model for natural language
  inference} (\bibinfo{year}{2016}).
\newblock \bibinfo{note}{Preprint at \url{https://arxiv.org/abs/1606.01933}}.

\bibitem{kim2017structured}
\bibinfo{author}{Kim, Y.}, \bibinfo{author}{Denton, C.},
  \bibinfo{author}{Hoang, L.} \& \bibinfo{author}{Rush, A.~M.}
\newblock \bibinfo{title}{Structured attention networks}
  (\bibinfo{year}{2017}).
\newblock \bibinfo{note}{Preprint at \url{https://arxiv.org/abs/1702.00887}}.

\bibitem{wang2020forward}
\bibinfo{author}{Wang, L.}, \bibinfo{author}{Zhang, C.}, \bibinfo{author}{Bai,
  R.}, \bibinfo{author}{Li, J.} \& \bibinfo{author}{Duan, H.}
\newblock \bibinfo{title}{Forward reaction prediction as reverse verification:
  A novel approach to retrosynthesis}.
\newblock \emph{\bibinfo{journal}{chemrxiv.13138775.v1}}
  (\bibinfo{year}{2020}).

\bibitem{bjerrum2017smiles}
\bibinfo{author}{Bjerrum, E.~J.}
\newblock \bibinfo{title}{Smiles enumeration as data augmentation for neural
  network modeling of molecules} (\bibinfo{year}{2017}).
\newblock \bibinfo{note}{Preprint at \url{http://arxiv.org/abs/1703.07076}}.

\bibitem{devlin2018bert}
\bibinfo{author}{Devlin, J.}, \bibinfo{author}{Chang, M.-W.},
  \bibinfo{author}{Lee, K.} \& \bibinfo{author}{Toutanova, K.}
\newblock \bibinfo{title}{Bert: Pre-training of deep bidirectional transformers
  for language understanding} (\bibinfo{year}{2018}).
\newblock \bibinfo{note}{Preprint at \url{https://arxiv.org/abs/1810.04805}}.

\bibitem{xue2021x}
\bibinfo{author}{Xue, D.} \emph{et~al.}
\newblock \bibinfo{title}{X-mol: large-scale pre-training for molecular
  understanding and diverse molecular analysis}.
\newblock \emph{\bibinfo{journal}{bioRxiv}} \bibinfo{pages}{2020--12}
  (\bibinfo{year}{2021}).

\bibitem{kishimoto2019depth}
\bibinfo{author}{Kishimoto, A.}, \bibinfo{author}{Buesser, B.},
  \bibinfo{author}{Chen, B.} \& \bibinfo{author}{Botea, A.}
\newblock \bibinfo{title}{Depth-first proof-number search with heuristic edge
  cost and application to chemical synthesis planning}.
\newblock \emph{\bibinfo{journal}{Advances in Neural Information Processing
  Systems}} \textbf{\bibinfo{volume}{32}} (\bibinfo{year}{2019}).

\bibitem{ishida2022ai}
\bibinfo{author}{Ishida, S.}, \bibinfo{author}{Terayama, K.},
  \bibinfo{author}{Kojima, R.}, \bibinfo{author}{Takasu, K.} \&
  \bibinfo{author}{Okuno, Y.}
\newblock \bibinfo{title}{Ai-driven synthetic route design incorporated with
  retrosynthesis knowledge}.
\newblock \emph{\bibinfo{journal}{Journal of chemical information and
  modeling}} \textbf{\bibinfo{volume}{62}}~(6), \bibinfo{pages}{1357--1367}
  (\bibinfo{year}{2022}).

\bibitem{chen2020retro}
\bibinfo{author}{Chen, B.}, \bibinfo{author}{Li, C.}, \bibinfo{author}{Dai, H.}
  \& \bibinfo{author}{Song, L.}
\newblock \bibinfo{title}{Retro*: learning retrosynthetic planning with neural
  guided a* search}.
\newblock In \emph{\bibinfo{booktitle}{International Conference on Machine
  Learning}} \bibinfo{pages}{1608--1616} (\bibinfo{organization}{PMLR},
  \bibinfo{year}{2020}).

\bibitem{xie2022retrograph}
\bibinfo{author}{Xie, S.} \emph{et~al.}
\newblock \bibinfo{title}{Retrograph: Retrosynthetic planning with graph
  search}.
\newblock In \emph{\bibinfo{booktitle}{Proceedings of the 28th ACM SIGKDD
  Conference on Knowledge Discovery and Data Mining}}
  \bibinfo{pages}{2120--2129} (\bibinfo{year}{2022}).

\bibitem{kim2021self}
\bibinfo{author}{Kim, J.}, \bibinfo{author}{Ahn, S.}, \bibinfo{author}{Lee, H.}
  \& \bibinfo{author}{Shin, J.}
\newblock \bibinfo{title}{Self-improved retrosynthetic planning}.
\newblock In \emph{\bibinfo{booktitle}{International Conference on Machine
  Learning}} \bibinfo{pages}{5486--5495} (\bibinfo{organization}{PMLR},
  \bibinfo{year}{2021}).

\bibitem{schwaller2020predicting}
\bibinfo{author}{Schwaller, P.} \emph{et~al.}
\newblock \bibinfo{title}{Predicting retrosynthetic pathways using
  transformer-based models and a hyper-graph exploration strategy}.
\newblock \emph{\bibinfo{journal}{Chemical science}}
  \textbf{\bibinfo{volume}{11}}~(12), \bibinfo{pages}{3316--3325}
  (\bibinfo{year}{2020}).

\bibitem{schreck2019learning}
\bibinfo{author}{Schreck, J.~S.}, \bibinfo{author}{Coley, C.~W.} \&
  \bibinfo{author}{Bishop, K.~J.}
\newblock \bibinfo{title}{Learning retrosynthetic planning through simulated
  experience}.
\newblock \emph{\bibinfo{journal}{ACS central science}}
  \textbf{\bibinfo{volume}{5}}~(6), \bibinfo{pages}{970--981}
  (\bibinfo{year}{2019}).

\bibitem{sumner2020levenshtein}
\bibinfo{author}{Sumner, D.}, \bibinfo{author}{He, J.},
  \bibinfo{author}{Thakkar, A.}, \bibinfo{author}{Engkvist, O.} \&
  \bibinfo{author}{Bjerrum, E.~J.}
\newblock \bibinfo{title}{Levenshtein augmentation improves performance of
  smiles based deep-learning synthesis prediction}.
\newblock \emph{\bibinfo{journal}{chemrxiv.12562121.v2}}
  (\bibinfo{year}{2020}).

\bibitem{carey2006analysis}
\bibinfo{author}{Carey, J.~S.}, \bibinfo{author}{Laffan, D.},
  \bibinfo{author}{Thomson, C.} \& \bibinfo{author}{Williams, M.~T.}
\newblock \bibinfo{title}{Analysis of the reactions used for the preparation of
  drug candidate molecules}.
\newblock \emph{\bibinfo{journal}{Organic \& biomolecular chemistry}}
  \textbf{\bibinfo{volume}{4}}~(12), \bibinfo{pages}{2337--2347}
  (\bibinfo{year}{2006}).

\bibitem{nicolaou1996classics}
\bibinfo{author}{Nicolaou, K.~C.} \& \bibinfo{author}{Sorensen, E.~J.}
\newblock \emph{\bibinfo{title}{Classics in total synthesis: targets,
  strategies, methods}}  (\bibinfo{publisher}{John Wiley \& Sons},
  \bibinfo{year}{1996}).

\bibitem{allis1994searching}
\bibinfo{author}{Allis, L.~V.} \emph{et~al.}
\newblock \emph{\bibinfo{title}{Searching for solutions in games and artificial
  intelligence}}  (\bibinfo{publisher}{Ponsen \& Looijen Wageningen},
  \bibinfo{year}{1994}).

\bibitem{hart1968formal}
\bibinfo{author}{Hart, P.~E.}, \bibinfo{author}{Nilsson, N.~J.} \&
  \bibinfo{author}{Raphael, B.}
\newblock \bibinfo{title}{A formal basis for the heuristic determination of
  minimum cost paths}.
\newblock \emph{\bibinfo{journal}{IEEE transactions on Systems Science and
  Cybernetics}} \textbf{\bibinfo{volume}{4}}~(2), \bibinfo{pages}{100--107}
  (\bibinfo{year}{1968}).

\bibitem{coley2018scscore}
\bibinfo{author}{Coley, C.~W.}, \bibinfo{author}{Rogers, L.},
  \bibinfo{author}{Green, W.~H.} \& \bibinfo{author}{Jensen, K.~F.}
\newblock \bibinfo{title}{Scscore: synthetic complexity learned from a reaction
  corpus}.
\newblock \emph{\bibinfo{journal}{Journal of chemical information and
  modeling}} \textbf{\bibinfo{volume}{58}}~(2), \bibinfo{pages}{252--261}
  (\bibinfo{year}{2018}).

\bibitem{li2017deep}
\bibinfo{author}{Li, Y.}
\newblock \bibinfo{title}{Deep reinforcement learning: An overview}
  (\bibinfo{year}{2017}).
\newblock \bibinfo{note}{Preprint at \url{https://arxiv.org/abs/1701.07274}}.

\bibitem{silver2017mastering}
\bibinfo{author}{Silver, D.} \emph{et~al.}
\newblock \bibinfo{title}{Mastering the game of go without human knowledge}.
\newblock \emph{\bibinfo{journal}{nature}}
  \textbf{\bibinfo{volume}{550}}~(7676), \bibinfo{pages}{354--359}
  (\bibinfo{year}{2017}).

\bibitem{mnih2015human}
\bibinfo{author}{Mnih, V.} \emph{et~al.}
\newblock \bibinfo{title}{Human-level control through deep reinforcement
  learning}.
\newblock \emph{\bibinfo{journal}{nature}}
  \textbf{\bibinfo{volume}{518}}~(7540), \bibinfo{pages}{529--533}
  (\bibinfo{year}{2015}).

\bibitem{silver2018general}
\bibinfo{author}{Silver, D.} \emph{et~al.}
\newblock \bibinfo{title}{A general reinforcement learning algorithm that
  masters chess, shogi, and go through self-play}.
\newblock \emph{\bibinfo{journal}{Science}}
  \textbf{\bibinfo{volume}{362}}~(6419), \bibinfo{pages}{1140--1144}
  (\bibinfo{year}{2018}).

\bibitem{schneider2016s}
\bibinfo{author}{Schneider, N.}, \bibinfo{author}{Stiefl, N.} \&
  \bibinfo{author}{Landrum, G.~A.}
\newblock \bibinfo{title}{What’s what: The (nearly) definitive guide to
  reaction role assignment}.
\newblock \emph{\bibinfo{journal}{Journal of chemical information and
  modeling}} \textbf{\bibinfo{volume}{56}}~(12), \bibinfo{pages}{2336--2346}
  (\bibinfo{year}{2016}).

\bibitem{jin2017predicting}
\bibinfo{author}{Jin, W.}, \bibinfo{author}{Coley, C.~W.},
  \bibinfo{author}{Barzilay, R.} \& \bibinfo{author}{Jaakkola, T.}
\newblock \bibinfo{title}{Predicting organic reaction outcomes with
  weisfeiler-lehman network} (\bibinfo{year}{2017}).
\newblock \bibinfo{note}{Preprint at \url{https://arxiv.org/abs/1709.04555}}.

\bibitem{lowe2012extraction}
\bibinfo{author}{Lowe, D.~M.}
\newblock \emph{\bibinfo{title}{Extraction of chemical structures and reactions
  from the literature}}.
\newblock Ph.D. thesis, \bibinfo{school}{University of Cambridge}
  (\bibinfo{year}{2012}).

\bibitem{schwaller2021mapping}
\bibinfo{author}{Schwaller, P.} \emph{et~al.}
\newblock \bibinfo{title}{Mapping the space of chemical reactions using
  attention-based neural networks}.
\newblock \emph{\bibinfo{journal}{Nature Machine Intelligence}}
  \textbf{\bibinfo{volume}{3}}~(2), \bibinfo{pages}{144--152}
  (\bibinfo{year}{2021}).

\bibitem{genheden2020aizynthfinder}
\bibinfo{author}{Genheden, S.} \emph{et~al.}
\newblock \bibinfo{title}{Aizynthfinder: a fast, robust and flexible
  open-source software for retrosynthetic planning}.
\newblock \emph{\bibinfo{journal}{Journal of cheminformatics}}
  \textbf{\bibinfo{volume}{12}}~(1), \bibinfo{pages}{1--9}
  (\bibinfo{year}{2020}).

\bibitem{irwin2020zinc20}
\bibinfo{author}{Irwin, J.~J.} \emph{et~al.}
\newblock \bibinfo{title}{Zinc20—a free ultralarge-scale chemical database
  for ligand discovery}.
\newblock \emph{\bibinfo{journal}{Journal of chemical information and
  modeling}} \textbf{\bibinfo{volume}{60}}~(12), \bibinfo{pages}{6065--6073}
  (\bibinfo{year}{2020}).

\bibitem{genheden2022paroutes}
\bibinfo{author}{Genheden, S.} \& \bibinfo{author}{Bjerrum, E.}
\newblock \bibinfo{title}{Paroutes: towards a framework for benchmarking
  retrosynthesis route predictions}.
\newblock \emph{\bibinfo{journal}{Digital Discovery}}
  \textbf{\bibinfo{volume}{1}}~(4), \bibinfo{pages}{527--539}
  (\bibinfo{year}{2022}).

\bibitem{he2020extended}
\bibinfo{author}{He, J.} \emph{et~al.}
\newblock \bibinfo{title}{An extended overview of the clef 2020 chemu lab:
  information extraction of chemical reactions from patents}.
\newblock In \emph{\bibinfo{booktitle}{Proceedings of CLEF (Conference and Labs
  of the Evaluation Forum) 2020 Working Notes}}  (\bibinfo{year}{2020}).

\bibitem{he2020overview}
\bibinfo{author}{He, J.} \emph{et~al.}
\newblock \bibinfo{title}{Overview of chemu 2020: named entity recognition and
  event extraction of chemical reactions from patents}.
\newblock In \emph{\bibinfo{booktitle}{International Conference of the
  Cross-Language Evaluation Forum for European Languages}}
  \bibinfo{pages}{237--254} (\bibinfo{organization}{Springer},
  \bibinfo{year}{2020}).

\bibitem{goodman2009computer}
\bibinfo{author}{Goodman, J.}
\newblock \bibinfo{title}{Computer software review: Reaxys}
  (\bibinfo{year}{2009}).

\bibitem{coley2019robotic}
\bibinfo{author}{Coley, C.~W.} \emph{et~al.}
\newblock \bibinfo{title}{A robotic platform for flow synthesis of organic
  compounds informed by ai planning}.
\newblock \emph{\bibinfo{journal}{Science}}
  \textbf{\bibinfo{volume}{365}}~(6453), \bibinfo{pages}{eaax1566}
  (\bibinfo{year}{2019}).

\bibitem{mikulak2020computational}
\bibinfo{author}{Mikulak-Klucznik, B.} \emph{et~al.}
\newblock \bibinfo{title}{Computational planning of the synthesis of complex
  natural products}.
\newblock \emph{\bibinfo{journal}{Nature}}
  \textbf{\bibinfo{volume}{588}}~(7836), \bibinfo{pages}{83--88}
  (\bibinfo{year}{2020}).

\bibitem{badowski2019selection}
\bibinfo{author}{Badowski, T.}, \bibinfo{author}{Molga, K.} \&
  \bibinfo{author}{Grzybowski, B.~A.}
\newblock \bibinfo{title}{Selection of cost-effective yet chemically diverse
  pathways from the networks of computer-generated retrosynthetic plans}.
\newblock \emph{\bibinfo{journal}{Chemical science}}
  \textbf{\bibinfo{volume}{10}}~(17), \bibinfo{pages}{4640--4651}
  (\bibinfo{year}{2019}).

\bibitem{kurczab2014influence}
\bibinfo{author}{Kurczab, R.}, \bibinfo{author}{Smusz, S.} \&
  \bibinfo{author}{Bojarski, A.~J.}
\newblock \bibinfo{title}{The influence of negative training set size on
  machine learning-based virtual screening}.
\newblock \emph{\bibinfo{journal}{Journal of cheminformatics}}
  \textbf{\bibinfo{volume}{6}}~(1), \bibinfo{pages}{1--9}
  (\bibinfo{year}{2014}).

\bibitem{beker2022machine}
\bibinfo{author}{Beker, W.} \emph{et~al.}
\newblock \bibinfo{title}{Machine learning may sometimes simply capture
  literature popularity trends: A case study of heterocyclic suzuki--miyaura
  coupling}.
\newblock \emph{\bibinfo{journal}{Journal of the American Chemical Society}}
  \textbf{\bibinfo{volume}{144}}~(11), \bibinfo{pages}{4819--4827}
  (\bibinfo{year}{2022}).

\bibitem{xu2017deep}
\bibinfo{author}{Xu, Y.}, \bibinfo{author}{Pei, J.} \& \bibinfo{author}{Lai,
  L.}
\newblock \bibinfo{title}{Deep learning based regression and multiclass models
  for acute oral toxicity prediction with automatic chemical feature
  extraction}.
\newblock \emph{\bibinfo{journal}{Journal of chemical information and
  modeling}} \textbf{\bibinfo{volume}{57}}~(11), \bibinfo{pages}{2672--2685}
  (\bibinfo{year}{2017}).

\bibitem{wu2021mining}
\bibinfo{author}{Wu, Z.} \emph{et~al.}
\newblock \bibinfo{title}{Mining toxicity information from large amounts of
  toxicity data}.
\newblock \emph{\bibinfo{journal}{Journal of Medicinal Chemistry}}
  \textbf{\bibinfo{volume}{64}}~(10), \bibinfo{pages}{6924--6936}
  (\bibinfo{year}{2021}).

\bibitem{jia2022explainable}
\bibinfo{author}{Jia, L.} \emph{et~al.}
\newblock \bibinfo{title}{Explainable fragment-based molecular property
  attribution}.
\newblock \emph{\bibinfo{journal}{Advanced Intelligent Systems}}
  \textbf{\bibinfo{volume}{4}}~(10), \bibinfo{pages}{2200104}
  (\bibinfo{year}{2022}).

\bibitem{murray2009rise}
\bibinfo{author}{Murray, C.~W.} \& \bibinfo{author}{Rees, D.~C.}
\newblock \bibinfo{title}{The rise of fragment-based drug discovery}.
\newblock \emph{\bibinfo{journal}{Nature chemistry}}
  \textbf{\bibinfo{volume}{1}}~(3), \bibinfo{pages}{187--192}
  (\bibinfo{year}{2009}).

\end{thebibliography}

\end{document}